%% file: arxiv_main.tex
\documentclass[11pt]{article}

\usepackage[letterpaper,margin=1in]{geometry}
\usepackage[parfill]{parskip}

\usepackage{authblk}

\usepackage[toc,page,header]{appendix}
\usepackage{natbib} % has a nice set of citation styles and commands
\usepackage{mathtools} % amsmath with fixes and additions

\usepackage{microtype}
\usepackage{graphicx}

\usepackage{subcaption}
\usepackage{booktabs} % for professional tables
\usepackage{multicol}
\usepackage{hyperref}
\usepackage{times}
\usepackage{appendix}
\usepackage{amsmath}
\usepackage{amsfonts}  
\usepackage{algorithmic,algorithm}

\usepackage{booktabs}
\usepackage{nicefrac} 
\usepackage{microtype}
\usepackage{xcolor}
\usepackage{multirow}
\usepackage{dsfont}
\usepackage{mathtools,enumitem}
\usepackage{tabularx}
\newcolumntype{C}{>{\centering\arraybackslash}X}
\usepackage{comment}
\usepackage{booktabs} % commands to create good-looking tables
\usepackage{tikz} % nice language for creating drawings and diagrams
\usepackage{amsmath}
\usepackage{amssymb}
\usepackage{amsthm}

\usepackage[capitalize,noabbrev]{cleveref}

\theoremstyle{plain}
\newtheorem{theorem}{Theorem}[section]

\newtheorem{lemma}[theorem]{Lemma}

\theoremstyle{definition}
\newtheorem{definition}[theorem]{Definition}

\theoremstyle{remark}
\newtheorem{remark}[theorem]{Remark}
\DeclareMathOperator*{\argmax}{arg\,max}

\def\cA{\mathcal{A}}

\def\cF{\mathcal{F}}

\def\cN{\mathcal{N}}
\def\cR{\mathcal{R}}

\def\cS{\mathcal{S}}

\def\cX{\mathcal{X}}
\def\cY{\mathcal{Y}}

\title{``Private Prediction Strikes Back!'' Private Kernelized Nearest Neighbors with Individual R\'{e}nyi Filter }
\author[1]{Yuqing Zhu}
\author[1]{Xuandong Zhao}
\author[2]{Chuan Guo}
\author[1]{Yu-Xiang Wang}

\affil[1]{%
    University of California, Santa Barbara
}
\affil[2]{%
    Facebook AI Research
}

\begin{document}
\maketitle

\input{01_abstract.tex}
\input{02_intro.tex}

\input{03_preliminary.tex}

\input{05_method.tex}

\input{06_experiment.tex}

\input{09_summary}

%\input{07_related.tex}

%\begin{contributions} % will be removed in pdf for initial submission 
%					  % (without ‘accepted’ option in \documentclass)
%                      % so you can already fill it to test with the
%                      % ‘accepted’ class option
%    Briefly list author contributions. 
%    This is a nice way of making clear who did what and to give proper credit.
%    This section is optional.
%
%    H.~Q.~Bovik conceived the idea and wrote the paper.
%    Coauthor One created the code.
%    Coauthor Two created the figures.
%\end{contributions}

\section*{Acknowledgements} % will be removed in pdf for initial submission,
						 % (without ‘accepted’ option in \documentclass)
                        % so you can already fill it to test with the
                        % ‘accepted’ class option
YZ, XZ and YXW are partially supported by NSF Award \#2048091. YZ was supported by a Google PhD Fellowship and XZ was supported by UCSB Chancellor's Fellowship.
\bibliography{egbib, DP}
\newpage
\appendix
\input{010_appendix}

\end{document}

%% file: 01_abstract.tex
\begin{abstract}

Most existing approaches of differentially private (DP) machine learning focus on \emph{private training}.  Despite its many advantages, \emph{private training} lacks the flexibility in adapting to incremental changes to the training dataset such as deletion requests from exercising GDPR’s \emph{right to be forgotten}. 
We revisit a long-forgotten alternative, known as \emph{private prediction} \citep{dwork2018privacy}, and propose a new algorithm named \emph{Individual Kernelized Nearest Neighbor} (Ind-KNN). Ind-KNN is easily updatable over dataset changes and it allows precise control of the R\'{e}nyi DP at an individual user level --- a user's privacy loss is measured by the exact amount of her contribution to predictions; and a user is removed if her prescribed privacy budget runs out. Our results show that Ind-KNN consistently improves the accuracy over existing private prediction methods for a wide range of $\epsilon$ on four vision and language tasks. We also illustrate several cases under which Ind-KNN is preferable over private training with NoisySGD. \footnote{code is available at \url{https://github.com/jeremy43/Ind_kNN}} 

\end{abstract}

%% file: 02_intro.tex
 % \vspace{-1em}
\section{Introduction}

Differential privacy (DP; \citet{dwork2006calibrating, dwork2014algorithmic}) is a promising approach for mitigating privacy risks in machine learning (ML). The predominant setting for private ML is to produce the model learned from sensitive data using DP primitives, a.k.a. private training ~\citep{chaudhuri2011differentially,kasiviswanathan2011can,abadi2016deep}. The resulting trained model can then be safely deployed with peace of mind, because DP ensures that no individual training sample can be identified from the model itself or its downstream predictions.

Unfortunately, private training comes with several irky properties that hamper its real-life deployment. 
To begin, private training comes at a significant computation cost that can be restrictive in many applications. The NoisySGD algorithm~\citep{abadi2016deep} requires per-sample gradient computation, which is much more computation- and memory-intensive than standard training.

Secondly, private training outputs a static model that cannot easily adapt to a changing dataset. For instance, additional data can arrive in a streaming fashion continuously. Also, training data could be mislabeled or corrupted~\citep{chen2017targeted, jagielski2018manipulating} and the model needs to be patched accordingly. In addition, if the model is trained on user data, privacy regulations such as GDPR entitle the user to request the removal of their data from the model~\citep{ginart2019making, guo2019certified, bourtoule2021machine} with the so-called \emph{right to be forgotten}~\citep{mantelero2013eu}. These requirements can be satisfied by periodically re-training the model, but such an approach is not applicable to private training due to its high computation cost as well as privacy degradation after repeated training runs.

Thirdly, privacy training operates under a very strong threat model in which all downstream users can collude with each other in a coordinated attack on any individual training sample. Sometimes it makes sense to make realistic assumptions that limit the adversaries' information or resources. For example, Harvard's Privacy Tools project (now OpenDP) adopts a weaker threat model where each downstream user keeps the results to themselves \citep{dptools}. In this way, they each get to spend the privacy budget independently of everyone else and enjoy higher utility. Private training unfortunately does not have a means to benefit from having weaker adversaries.

\begin{table*}[ht]
\centering
\caption{ The amortized computational and privacy cost of answering $T=2000$ queries on CIFAR-10. The median accuracy of all approaches across five independent runs is aligned to $96.0\%$. We estimate the amortized computational cost by calculating the averaged time spent (in seconds) to answer a single query, which is the total time of training divided by $T$ in Linear NoisySGD~\citep{feldman2021individual} and the total time of predictions divided by $T$ in Private kNN~\citep{zhu2020private} and Ind-KNN. We use $\delta=10^{-5}$.  In the retraining scenarios, we assume that a retraining request is made every answering 100 queries, resulting in a total of 20 retraining requests among $T$ queries.}
% \vspace{-0.5em}
\resizebox{\textwidth}{!}{
\begin{tabular}{c|c|c|c|c|c}
    \toprule
    &   NoisySGD & NoisySGD (with retrain) & Private kNN & Ind-KNN (ours) & Ind-KNN+hashing (ours)  \\
    \hline
     Computational cost (s)  & 0.008 &0.16 &0.12 & 0.25 & 0.04 \\ 
      \hline
      Privacy loss ($\epsilon$)   &  1.5 & 6.2 & 4.1 & 2.0  & 3.2\\ 
    \hline
\end{tabular}}
\label{tab: intro}
 \vspace{-0.5em}
\end{table*}

To address these issues of private training, we revisit a viable but less-known alternative setting in differentially private machine learning known as \emph{privacy-preserving prediction} (or simply \emph{private prediction}) \citep{dwork2018privacy}. Instead of privately training the models and then using the model for predictions, private prediction aims at generating a sequence of predictions using the data directly. Notable methods include those that perturb the predictions of non-private models \citep{dwork2018privacy,pate2018,bassily2018,dagan2020pac}, or those that perturb the voting scores of the nearest neighbors \citep{zhu2020private}. These methods require no changes to the (non-private) data workflow, and thus could more easily adapt to changing data. 

%Readers who are familiar with DP basics may complain \textit{``Wait a minute! An ML model needs to serve many queries. Isn't the privacy-utility tradeoff going down the toilet?''} 
 
From the privacy-utility trade-off point of view, the private prediction setting may appear to be counter-intuitive, because, for every prediction that it generates, a unit of privacy budget is spent. It is unreasonable to expect private prediction methods to outperform private training methods such as NoisySGD when we need to make many predictions. This was well-documented in the work of \citet{van2020trade}.  However, in the aforementioned situations when either frequent data updates are needed or a weaker adversary is assumed\footnote{Consider the example of a recommendation system, each user makes a much smaller number of predictions than all users collectively.}, private prediction methods can significantly outperform NoisySGD (See Table~\ref{tab: intro} and Figure~\ref{fig: unlearn} for an illustration).  In fact, we will demonstrate that when combined with modern DP accounting techniques, data-adaptive DP algorithm design, and some clever reuse of previous predictions, a small privacy budget can answer thousands of queries without significantly increasing the privacy loss.

In this work, we propose \emph{Individual Kernelized Nearest Neighbors} (Ind-KNN) --- a new private prediction mechanism that significantly increases the number of queries one can answer with an individualized R\'{e}nyi Differential Privacy accountant and other techniques. 
Intuitively, in KNN prediction, training samples that do not belong to the query's neighbor set do not contribute to the prediction, and hence their privacy cost should be negligible.
We show that by slightly modifying KNN and leveraging R\'{e}nyi filter~\citep{feldman2021individual} to account for the privacy cost of each sample individually, we can realize this intuition in the privacy accounting and allow each training sample to participate in the query response until its own privacy budget is exhausted.
In effect, common queries can be answered with relatively low privacy costs due to a large number of similar samples present in the training set.

% \paragraph{Contributions} Our contributions are the following:
\textbf{Experimental results.} We summarize our experimental results as follows:
\begin{enumerate}
 %\vspace{-1em}
\item We show that Ind-KNN consistently outperforms the private prediction benchmark, Private-kNN~\citep{zhu2020private}, across four vision and language tasks for a range of epsilon between $[0.5, 2.0]$. 
\item We demonstrate that Ind-KNN is a viable alternative to private training methods even in a static data setting. Our results indicate that Ind-KNN achieves higher accuracy than NoisySGD when answering less than 2000 queries on CIFAR-10 under $(1.0, 10^{-5})$-DP. 
\item For frequent data updates, Ind-KNN significantly outperforms the private training benchmark Linear NoisySGD~\citep{feldman2021individual}. As shown in Table~\ref{tab: intro}, Linear NoisySGD requires a DP budget of $\epsilon=6.2$ to achieve an accuracy of $96.0\%$ on $2000$ queries of CIFAR-10, while Ind-KNN only requires $\epsilon=2.0$.

\item We describe two simple techniques that significantly enhance the computational efficiency and utility of Ind-KNN. First, we show  that incorporating hashing tricks into Ind-KNN can provide a $6\times$ speedup in making predictions, with only a negligible drop in accuracy. Additionally, we propose to reuse the results of previous queries via post-processing, which allows Ind-KNN to answer an additional $1000$ queries on CIFAR-10 without compromising in privacy or utility.
%\item We also propose two techniques to enhance Ind-KNN's efficiency and utility: incorporating hashing tricks for a $6\times$ speedup with only a negligible drop in accuracy and reusing predictions for an additional 1000 queries on CIFAR-10 without compromising privacy or utility.
 % \vspace{-0.8em}
\end{enumerate}

\noindent\textbf{Related work and novelty.} 
The problem of private prediction was pioneered by \citet{dwork2018privacy} as a weakened goal for private machine learning. Model-based approaches for private predictions either require analyzing the  stability of model training \citep{dwork2018privacy,dagan2020pac} or to enforce stability of prediction via subsample-and-aggregate \citep{pate2018,bassily2018}. Our method is closest to Private kNN \citep{zhu2020private} but uses kernel-weighted neighbors with a variable $K$ instead of a fixed $K$. This change is critical for adapting the individual R\'{e}nyi DP accountant (and filter) for our purpose. Other components such as adaptive noise-level, prediction reuse, and the fast hashing trick are new to this paper. Technically, we apply the same individual R\'{e}nyi filter \citep{feldman2021individual} that retires data samples when their privacy budget runs out. The difference is that we applied it to KNN rather than noisy gradient descent. KNN naturally has bounded support thus it is efficient to maintain the individual RDP accountants.

%% file: 03_preliminary.tex
 % \vspace{-1em}
\section{Preliminaries}\label{sec:preliminary}

We start with the definition of differential privacy.
\begin{definition}[Differential Privacy~\citep{dwork2006calibrating}] \label{def:dp}
	A randomized algorithm $\cA : \cX \to \Theta$ is $(\epsilon,\delta)$-DP (differentially private) if for every pair of neighboring datasets $S, S'\in \cS$, and every possible (measurable) output set $E \subseteq \Theta$ the following inequality holds: $$\Pr[\cA(S) \in E] \leq e^{\epsilon} \Pr[\cA(S') \in E] + \delta.$$
\end{definition} 

\begin{definition}[R\'enyi Differential Privacy \citep{mironov2017renyi}] \label{def:RDP}
	We say that a mechanism $\cA$ is $(\alpha, \epsilon(\alpha))$-RDP with order $\alpha\in (1,\infty)$ if for all neighboring datasets $S,S'$: 
	\begin{align*}
	&D_{\alpha}(\cA(S)\|\cA(S') )\\
	&= \frac{1}{\alpha-1}\log E_{\theta\sim \cA(S')}\left[ \left(\frac{p_{\cA(S)}(\theta)}{p_{\cA(S')}(\theta)}\right)^\alpha \right] \leq \epsilon(\alpha).
	\end{align*}
\end{definition}
As $\alpha\rightarrow \infty$, RDP converges to the standard $(\epsilon,0)$-DP. More generally, we can convert RDP to standard $(\epsilon, \delta)$-DP for any $\delta>0$ using conversions from 
~\citep{balle2020hypothesis}.

\noindent\textbf{Privacy composition.} RDP features a natural composition theorem that significantly simplifies privacy analysis over compositions, and often leads to a tighter privacy guarantee.   If $\cA_1(\cdot)$ is $(\alpha, \epsilon_{\cA_1}(\alpha))$-RDP and $\cA_2(\cdot)$ is $(\alpha, \epsilon_{\cA_2}(\alpha))$-RDP, then the adaptive composition theorem for RDP says that $\epsilon_{\cA_1 \circ \cA_2}(\cdot)$ satisfies $(\alpha, \epsilon_{\cA_1}(\alpha)+\epsilon_{\cA_2}(\alpha))$-RDP.

\noindent\textbf{Privacy-preserving prediction.} 
% \paragraph{Privacy-preserving prediction}
We now formally state the setting of privacy-preserving prediction. Consider a prediction task over a domain $\cX$ and label space $\cY$.  The prediction interface $\cA$ has access to a 
private dataset $S=(x_i, y_i)_{i=1}^n \in (\cX \times \cY)^n$, which outputs a value $a \in \cY$ if given a query $q \in \cX$. We denote by $Q$ a query generating algorithm that can adaptively generate a query given the previous released outputs.  Namely, we denote by  $\cA(S)  \rightleftharpoons_T Q =(q_t, a_t)_{t=1}^T$ the sequence of query-response pairs generated by the prediction interface $\cA$ over a sequence of length $T$ queries on dataset $S$, where $a_t=\cA_t(a_1, ..., a_{t-1}, S, q_t)$.

The privacy guarantee of private prediction is applied for a sequence of predictions generated by the interface $\cA$.

\begin{definition}[Privacy-preserving prediction interface]\citep{dwork2018privacy}
 A prediction interface $\cA$ is $(\epsilon, \delta)$-differentially private, if for every interactive query generating algorithm $Q$, the output $\cA(S)  \rightleftharpoons_T Q =(q_t, a_t)_{t=1}^T$ is $(\epsilon, \delta)$-DP with respect to dataset $S$.

\end{definition}

Privacy-preserving prediction algorithms can be useful in a variety of situations where releasing a DP model is restricted or not practical. 
For example, companies that train a privacy-preserving model and only require making a limited number of predictions can rely on a prediction interface instead of releasing the entire model. In addition, in health or financial data scenarios, private prediction algorithms allow for a cloud-based interface to be exposed, which can also help to ensure compliance with regulatory requirements.

\noindent\textbf{Individual RDP.}
% \paragraph{Individual RDP}
Our privacy analysis relies on individual privacy loss, which accounts for the maximum possible impact of an individual data point on a dataset. The following definition states the individual privacy loss in terms of R\'{e}nyi divergence. 
\begin{definition}[Individual RDP~\citep{feldman2021individual}]\label{def: indRDP}
Fix $n\in \cN$ and a private data point $z=(x, y)\in \cX\times\cY$. We say that a randomized algorithm $\cA$ satisfies $(\alpha, \rho)$-individual R\'{e}nyi differential privacy for $z$ if for all datasets $S=(z_1, ..., z_m)$ such that $m\leq n$ and $z_i=z$ for some $i$, it holds that
\[
D_\alpha^{\leftrightarrow}\big(\cA(S)||\cA(S^{-i})\big)\leq \rho,
\] where $D_\alpha^\leftrightarrow$ denotes the max of $D_{\alpha}\big(\cA(S)\|\cA(S^{-i})\big)$ and $D_{\alpha}\big(\cA(S^{-i})\|\cA(S)\big)$.
\end{definition}
%We use the $\leftrightarrow$ notation to denote the two directions of R\'{e}nyi divergence.

Note that the individual RDP parameter $\rho$ is a function of a data point $z$, and thus does not imply the standard RDP guarantee in Definition \ref{def:RDP}. However, we can obtain the standard RDP guarantee by requiring that all data points $z$ satisfy individual RDP with the same $\rho$.

Now, we present an example of individual RDP computation on Gaussian mechanism.
\begin{lemma}[Linear queries with Gaussian mechanism~\citep{feldman2021individual}]\label{lem: ind_linear}
Let $S=(z_1, ..., z_n)\in (\cX \times \cY)^n$. Suppose that $\cA$ is a $d$-dimensional linear query with Gaussian noise addition, $\cA(S)=\sum_{j\in[n]}q(z_j) + \cN(0,\sigma^2 \mathbf{1}_d)$ for some $q:\cX\times \cY \to \cR^d$. Then $\cA$ satisfies
\[D_\alpha^{\leftrightarrow}\big(\cA(S)||\cA(S^{-i})\big)\leq \frac{\alpha ||q(z_i)||_2^2}{2\sigma^2}\]
individual $RDP$ for $z_i$. Note that by replacing$||q(z_i)||_2$ with the $\ell_2$ global sensitivity of $q(\cdot)$, the expression above recovers the standard RDP of Gaussian mechanism.
\end{lemma}

% why needs fully adaptive composition?

The following theorem states the composition property of individual privacy. For a sequence of algorithms, as long as the composition of individual RDP parameters does not exceed a pre-specified budget for all data points, the output of the adaptive composition preserves the standard RDP guarantee.
\begin{theorem}[Corollary 3.3~\citep{feldman2021individual}]\label{thm: comp}
Fix any $G\geq 0$ and any $\alpha\geq 1$. For any input dataset $S=(z_1, ..., z_n)$ and for any sequence of algorithms $\cA_1, ..., \cA_T$, let $\rho_t^{(i)}$ denote the individual RDP parameter of the $t$-th adaptively composed algorithm $\cA_t$ with respect to $z_i$.
if  $\sum_{t=1}^T\rho_t^{(i)} \leq G$ holds almost surely for all $i\in[n]$ then the adaptive composition $\cA^{(T)}$ satisfies $(\alpha, G)$-RDP.
\end{theorem}
The composition rule described above is known as fully adaptive composition~\citep{rogers2016privacy}, which takes \textit{adaptively-chosen} privacy parameters instead of pre-specified ones in the classical adaptive composition. This type of composition is necessary for individual privacy since the individual RDP parameters themselves are random variables that depend on the outputs released by previous composed mechanisms.

To implement the composition above, we need a tool called \emph{R\'{e}nyi filter}, which is designed to ensure that the composed individual privacy parameters is maintained within a given budget $G$ for all individuals.
% provide an explicit example of Renyi filter.
In practice, we can implement R\'{e}nyi filter by providing each data point with an individual accountant that estimates its composed individual RDP $\sum_{t=1}^T \rho_t^{(i)}$ and dropping the data point once it exceeds the budget, as shown in Algorithm~\ref{alg: filter}. 

However, despite its tighter privacy analysis, this technique has been criticized for its computational cost of tracking individual privacy costs for all data samples. In this work, we demonstrate that KNN works seamlessly with the individual RDP accountant. Only selected neighbors are required to update their individual privacy accountants, which significantly reduces the computational cost.

\begin{algorithm}[htbp]
 \caption{Adaptive composition $\cA^{(T)}$ with R\'enyi filter}
 \label{alg: filter}
 \begin{algorithmic}[1]
 \STATE{\textbf{Input}: Dataset $S\in (\cX \times \cY)^n$, sequence of algorithms $\cA_{1:T}$ and privacy budget $G$}.
 \FOR{$t=1, ..., T$}
 \STATE{For all $z_i \in S$, compute \\\begin{small}$\rho_t^{(i)}=\sup_{S'\in \cS}D_{\alpha}^{\leftrightarrow}\left(\cA(a_{1:t-1}, S')||\cA(a_{1:t-1}, S'^{-i})\right)$\end{small}}
 \STATE{Update the active set $S=\{z_i| \sum_{j=1}^t \rho_j^{(i)}\leq G\}$}
 \STATE{Compute $a_t=\cA_t(a_{1:t-1}, S)$}
 \ENDFOR
 \STATE{\textbf{Return} $(a_1, ..., a_T)$}
 \end{algorithmic}
 \end{algorithm}

%% file: 05_method.tex
 % \vspace{-1em}
\section{Private Prediction with Ind-KNN}
 % \vspace{-0.5em}
%\section{Private Prediction with Ind-KNN}
\label{sec:main}

To overcome the limitations of private training, we propose \emph{Individual Kernelized Nearest Neighbor} (Ind-KNN)---a k-nearest neighbor-based private prediction algorithm that achieves a comparable DP guarantee and test accuracy to that of private training.

\noindent\textbf{Notations and setup.} 
% \paragraph{Notations and setup}
We focus on the task of multi-class classification. 
Given a private dataset $S=(x_i, y_i)_{i=1}^n \in (\cX \times \cY)$, we assume $y_i$ is an one-hot vector over $c$ class, i.e., $y_i \in\{0,1\}^c$. 
Let $\phi(\cdot)$ denote a \emph{public} feature extractor that maps the input $x\in \cX$ to a fixed-length feature representation $\phi(x)\in \cR^d$. This could be image features extracted from the penultimate layer of a ResNet50 pre-trained model or language features extracted from the final layer of a transformer model. The feature extractor is used to encode both the private dataset and public queries.
% need a description of the private dataset
\begin{algorithm}[htbp]
\caption{Privacy-preserving prediction with naive kNN}
\label{alg: naive_ind_kNN}
\begin{algorithmic}[1]
\STATE{\textbf{Input}: Dataset $S\in (\cX \times \cY)^n$, sequence of queries $q_1, ..., q_T$, number of neighbor $k$ and the noisy scale $\sigma$.}
\FOR{$t=1$ to $T$}
\STATE{$\cN_k:=$ top k nearest neighbors of the query $q_t$}
\STATE{$a_t = \argmax_{j\in[c]}\left(\sum_{i \in \cN_k }y_i +\cN(\boldsymbol{0}, \sigma^2\mathbf{1}_c) \right)_j$}
\ENDFOR
\STATE{\textbf{Return} $(a_1, ..., a_T)$}
\end{algorithmic}
\vspace{-1mm}
\end{algorithm}

% why knn does not work?
% why need to release k?
% why using kernel?
Previously, k-Nearest Neighbor (kNN) has been used for privacy-preserving prediction by \citet{zhu2020private} (Algorithm \ref{alg: naive_ind_kNN}).
In this method, when a query $q_t$ arrives, the top k nearest neighbors are selected from the private dataset based on the distance in the feature space, and their labels are utilized for prediction through a Gaussian mechanism.

However, the privacy loss of Algorithm~\ref{alg: naive_ind_kNN} 
accumulates rapidly as the number of queries increases, owing to its conservative privacy analysis that bounds the worst-case individual privacy loss over all individuals. In contrast, the Ind-KNN approach emphasizes individual privacy accounting, providing precise control over privacy loss at an individual data level. This allows each data point's privacy to be charged by the exact amount of its contribution to the query response, and private data is removed once its own privacy budget has been exhausted.

\begin{algorithm}[htbp]
\caption{Kernelized-nearest-neighbor with individual privacy accounting (Ind-KNN)}
\label{alg:ind_kNN}
\begin{algorithmic}[1]
\STATE{\textbf{Input}: Dataset $S\in (\cX \times \cY)^n$, the kernel function $\kappa(\cdot, \cdot)$, the threshold $\tau$, sequence of queries $q_{1:T}$, the noisy scale $\sigma_1$, $\sigma_2$ and the individual budget $B$. }
\STATE{Initialize individual budget $z_i = B, \forall i \in [n]$.}
\FOR{$t=1$ to $T$}
\STATE{Update the active set $S=\{(x_i, y_i)|z_i \geq \frac{1}{2\sigma_1^2}\}$.}
\STATE{Release the number of selected neighbors: $K_t:= \sum_{(x_i, y_i) \in S} \mathbb{I}[\kappa(x_i, q_t)\geq \tau] + \cN(0, \sigma_1^2).$}
\FOR{$(x_i, y_i)\in S$}
\STATE{Update the remaining budget $z_i$ after releasing $K_t$: $z_i=z_i - \frac{1}{2\sigma_1^2}\cdot \mathbb{I}[\kappa(x_i, q_t)\geq \tau]$. }
\STATE{Evaluate individual contribution $f_t: \cX \times \cY \to \cR^c$ as $f_t(x_i, y_i):=\min\big(\kappa(x_i, q_t)\cdot y_i \cdot \mathbb{I}[\kappa(x_i, q_t)\geq \tau], \sigma_2\sqrt{2K_t\cdot z_i}\cdot \mathbf{1}_c\big)$}

\STATE{Update the remaining budget $z_i$ after releasing label: $z_i = z_i - \frac{||f_t(x_i, y_i)||_2^2}{2\sigma_2^2\cdot K_t}$.}
\ENDFOR
\STATE{$a_t = \argmax_{j \in [c]}\big(\sum_{(x_i, y_i)\in S} f_t(x_i, y_i)+\cN(\boldsymbol{0}, \sigma_2^2\cdot K_t\cdot\mathbf{1}_c)\big)_j$.}
%\STATE{$a_t$ = Algorithm~\ref{alg:noisy_label}($S, \kappa(\cdot, \cdot), q_t, \sigma_1, \sigma_2$).}
\ENDFOR
\STATE{\textbf{Return} $(a_1, ..., a_T)$}
\end{algorithmic}
\vspace{-1mm}
\end{algorithm}

We propose a novel solution \emph{Individualized Kernelized Nearest Neighbor} (Ind-KNN) in Algorithm~\ref{alg:ind_kNN}. Intuitively, nearest neighbor-based prediction leak little to no private information when the query point is near a dense region of the training data. This is because the result of the query is determined by a large number of training samples and hence is insensitive to individual training points. We make several modifications to Private kNN to realize this intuition.

First, we introduce individual privacy accounting by assigning each private data point $(x_i, y_i)$ with a pre-determined privacy budget $B$, represented by the variable $z_i:=B$. For each query, the algorithm updates the private dataset $S$ to only include data points where $z_i>\frac{1}{2\sigma_1^2}$. This ensures that the privacy budget for each individual is not exceeded. 

Second, Ind-KNN improves upon Algorithm~\ref{alg: naive_ind_kNN} by utilizing a pre-specified threshold $\tau$ and a kernel-based similarity function $\kappa(\cdot, \cdot)$ to select only neighbors with similarity above $\tau$. This approach allows only the selected neighbors to be accountable for their privacy loss, preserving the privacy budget of un-selected private individuals for future queries. It is worth noting that simply selecting the exact top k neighbors, as in Algorithm~\ref{alg: naive_ind_kNN}, is not consistent with individual privacy loss. This is because the decision of selection is dependent on the dataset: a $k+1$th nearest neighbor in one dataset may be the top nearest neighbor in another dataset. Hence, all private data points must account for their individual privacy loss, even if only a subset of them contribute to the prediction, according to the definition of individual RDP (Definition~\ref{def: indRDP}).

Moreover, Ind-KNN employs kernel weights for prediction aggregation instead of equal weight for all nearest neighbors. In our experiments, we consider two types of kernel functions, RBF and cosine, to measure the similarity. For example,  the RBF kernel is defined as $\kappa(x, q_t):= e^{\frac{-||\phi(x)-\phi(q_t)||_2^2}{\nu^2}}$, where $\phi(x)$ and $\phi(q_t)$ are the encoded feature and $\nu$ is a scalar parameter. This adaptation, made possible by individual privacy accounting, results in a more accurate characterization of each individual's contribution to the query. However, changing from equal weight to kernel weight in Algorithm~\ref{alg: naive_ind_kNN} would not alter its privacy analysis (as the worst-case kernel weight is bound by 1), but would instead decrease the signal-to-noise ratio (each neighbor's contribution would be less than 1).

Finally, Ind-KNN dynamically adjusts the magnitude of noise added to the noisy prediction by publishing the number of neighbors at each query.
We find that adding noise with variance proportional to $K_t$ is crucial for good performance. This allows us to adjust the margin of the voting space --- the difference between the largest and the second largest coordinate of  $\sum_{(x_i, y_i)\in S} f(x_i, y_i)$ adapted to the noise scale. Specifically, when the margin is significant, adding larger noise will not change the output label, but it reduces each individual's individual privacy loss proportional to the reciprocal of $K_t$, enabling them to participate in more queries in the future.

\textbf{Algorithm.} The modifications made in Ind-KNN are summarized in Algorithm~\ref{alg:ind_kNN}. Specifically, since the number of selected neighbors is considered private information, each selected neighbor accounts for its individual  privacy loss due to releasing $\mathbb{I}[\kappa(x_i, q_t)\geq \tau]$ by subtracting $z_i$ with $\frac{1}{2\sigma_1^2}$ at line 7 of Algorithm~\ref{alg:ind_kNN}. Meanwhile, 
$f_t(x_i, y_i)$ at line 8 accounts for the individual contribution of releasing its label associated with kernel weight. 
The first term represents the ``weighted'' one-hot label for selected neighbors and all-zero vectors for unselected private data.
The second term $ \sigma_2\sqrt{2 K_t z_i}$ ensures that the incurred individual privacy loss of releasing label will not go beyond the remaining budget $z_i$.

\begin{lemma}[Individual RDP of releasing $a_t$]
Given a query $q_t$, for each $(x_i, y_i)$, define the function $f_t: \cX \times \cY \to \cR^c $ as line $8$ in Algorithm~\ref{alg:ind_kNN}.  Then the release of $a_t = \argmax_{j\in [c]}\big(\sum_{(x_i, y_i)\in S}f_t(x_i, y_i) +\cN(0, \sigma_2^2 \cdot K_t \cdot \mathbf{I}_c)\big)_j$ satisfies $(\alpha, \frac{\alpha \cdot ||f_t(x_i, y_i)||_2^2}{2\sigma_2^2\cdot K_t})$ individual RDP for each $(x_i, y_i)$.
\end{lemma}
The proof directly follows from Lemma~\ref{lem: ind_linear} and the post-processing property of individual privacy. Note that for unselected private data, their individual privacy loss is always zero since their individual contribution $f(\cdot)$ is zero.
\begin{theorem}\label{thm: privacy_indknn}
Algorithm~\ref{alg:ind_kNN} satisfies $(\alpha, B\alpha)$-RDP for all $\alpha\geq 1$.
\end{theorem}
\begin{proof}[Proof sketch]
The proof (deferred to the appendix) makes use of the facts that: (1) the decision rule for ``being selected''  is not influenced by any other private data points, thus, ``unselected'' neighbors does not incur any individual privacy loss. (2) adding/removing one selected neighbor would only change $\sum_{(x_i, y_i)\in S} \mathbb{I}[\kappa(x_i, q_t)\geq \tau]$ by one, thus the release of $K_t$ satisfies $(\alpha, \frac{\alpha}{2\sigma_1^2})$ individual RDP for selected neighbors. (3)  the release of label associated with the kernel weight satisfies $(\alpha, \frac{\alpha ||f_t(x_i,y_i)||_2^2}{2\sigma_2^2 K_t})$ individual RDP.
\end{proof}
\begin{remark}
We remark that the privacy guarantee of Ind-KNN is determined by the given individual budget, and remains the same regardless of the number of predictions made. However, as the number of predictions increase, the exclusion of private data may result in a degradation of the algorithm's utility.
\end{remark}

% add a remark 
% \vspace{-1.em}
\subsection{Efficient Ind-KNN}\label{sec: hash}
% \vspace{-0.5em}
In this section, we present two novel techniques that aim to improve the efficiency of Ind-KNN in terms of both utility and computational cost.

% in which scenarios we can reuse prediction?
% \paragraph{Ind-KNN with prediction reuse}
\noindent\textbf{Ind-KNN with prediction reuse.}
The first technique improves the utility of Ind-KNN by exploiting the previously released predictions. We acknowledge that the query-response pairs that have been disclosed can be considered public information. Therefore, we incorporate those predictions into the active set $S$ without any limitation on their privacy budgets. The results of our experiments demonstrate that this extension effectively mitigates the utility loss caused by the exclusion of private data points and improves the test accuracy when handling a large number of queries.

% \paragraph{Ind-KNN with hashing}
\noindent\textbf{Ind-KNN with hashing.}
Algorithm~\ref{alg:ind_kNN} requires searching through all private data to answer each query, which can be computationally expensive if the private dataset is large.
To address this issue, we present a variant of Ind-KNN that incorporates locality-sensitive hashing (LSH)~\citep{gionis1999similarity} for efficient nearest neighbor search. The full algorithm of Ind-KNN-Hash is in the appendix.
LSH is a well-established technique to speed up the approximate nearest neighbor search.  The principle behind the algorithm is to apply LSH to group private data points into ``buckets'' based on their hash values. When a query is made, the algorithm only needs to search the bucket that the query falls into, rather than searching through the entire dataset. 

Concretely, Ind-KNN-Hash creates $L$ hash tables $\cF=(f_1, ..., f_L)$ with each of them maps a feature $\mu\in \cR^d$ to a $b$-dimension bucket.  For each table $f$, the algorithm generates $b$ independent random Gaussian vectors from $\cN(\boldsymbol{0}, \mathbf{1}_d)$, denoted by $r_j$ for $1\leq j\leq b$. Then we encode $\mu$ with $f(\mu) = (h_1(\mu),..., h_b(\mu))$, where $h_j(\mu)=0$ if $r_j^\top \mu <0$, otherwise $h_j(\mu)=1$. The algorithm then indexes all private data points into the hash tables using their encoded features. When a query $q_t$ is received, the algorithm uses LSH to retrieve a set of private data points that are hashed into the same bucket in at least one table, which is denoted by $\cF(q_t)$. Finally,  Algorithm~\ref{alg:ind_kNN} is called to label each query with a slight modification on the active set, which is now restricted to the retrieved data points with non-negative individual budgets. Typically, increasing the number of hash tables $L$ and reducing the bucket size $b$ results in more accurate neighbors but higher computational costs. 

Incorporating LSH into Ind-KNN does not impose any additional privacy cost. This is because the encoding of each private data point is based on random Gaussian vectors and is executed independently of any other private data points.

%% file: 06_experiment.tex
\section{Experiments}\label{sec:exp}
 % \vspace{-1em}
%\subsection{Datasets and Features}\label{sec:feature}
We consider the following standard image classification and language classification datasets. For each dataset, we take the training set as the private domain and the testing set as the public domain.% We report the averaged prediction accuracy on a random subset from the testing set. 

\noindent\textbf{Image classification.} 
% \paragraph{Image classification}
We evaluate our method on two widely used image classification benchmarks, CIFAR-10 \citep{Krizhevsky2009LearningML} and Fashion MNIST \citep{Xiao2017FashionMNISTAN}. For CIFAR-10, we employed the recent Vision Transformer (ViT) model~\citep{vit}, which is pre-trained on the ImageNet-21k consisting of 14 million images and 21843 classes). The extracted from the ViT model are represented as 768-dimensional vectors. For Fashion MNIST, we consider the publicly available ImageNet-pretrained ResNet50 \cite{He_2016_CVPR} from Pytorch as the feature extractor. The model returns a $1000$-dim vector for each input image.

\noindent\textbf{Text classification.} 
We utilize AG News \citep{Zhang2015CharacterlevelCN} and DBPedia \citep{Lehmann2015DBpediaA} datasets to evaluate the performance of Ind-KNN on text classification tasks. We employ sentence embedding models \citep{zhao-etal-2022-compressing, reimers-2019-sentence-bert} to extract features. Specifically, we utilize the \texttt{all-roberta-large-v1} sentence-transformer, which has been fine-tuned on a 1B sentence pairs dataset using a self-supervised contrastive learning objective. The extracted features are 1024-dimensional vectors for each text instance.

% \begin{enumerate}
% \item \textbf{AG News} is a 4-class text classification dataset that consists of $120k$ training data and $7600$ testing data.
% \item \textbf{DBPedia} instances were selected from the English, non-regional Extended Abstracts provided by the DBpedia site, forming a 14-class text classification task. We utilize sentence-transformer to extract the feature in AGNews and DBPedia dataset.
% \end{enumerate}
%We evaluate the privacy and utility of $2000$ queries. We use a ResNet-50 model pre-trained on ImageNet to extract features for all methods. The feature dimension is $1000$.
%\subsection{Baselines}
We consider the following two algorithms for comparisons:

 \textbf{Linear+NoisySGD}~\citep{tramer2020differentially} is a private training benchmark that has been shown outperforming end-to-end privacy-preserving deep learning methods (including those pre-trained on public data, see \citet{de2022unlocking}) for a wide range of $\epsilon$. We consider this algorithm as a reference point for private training to investigate how well Ind-KNN performs compared to private training while we gain those computational savings. We implement the algorithm by training a linear model with features extracted from the same extractor as Ind-KNN. We use the default batch size $256$ and clip the gradient norm to $0.1$. The model is trained for 10 epochs with a grid search over the learning rate and the noise level is determined by the target privacy budget.
    
 \textbf{Private-kNN}~\citep{zhu2020private} is a private prediction baseline that we consider. For each query, the algorithm first samples a random subset from the private dataset, retrieves the k-nearest neighbors from the subset (based on the extracted features), and then releases the noisy label of kNN prediction using Report-Noisy-Max. We tune the sampling ratio and the number of neighbors on the validation set. The noise scale is calibrated based on the target privacy budget.

%Figure plots t
% \begin{table*}[ht]
% \centering
% \caption{Median accuracy on CIFAR-10 under $(2.0, 10^{-5})$-DP. }
% \resizebox{1.9\columnwidth}{!}{
%     \begin{tabular}{c@{\hskip 5mm}c@{\hskip 5mm}ccc}
%         \toprule
%         Methods & \#  Acc ($\%$) on $400$ queries & Acc ($\%$) on $800$ queries  \\
%         \hline
%          Linear+NoisySGD\citep{tramer2020differentially}      & $83.4$  & $83.4$  \\ Private kNN\citep{zhu2020private} &$82.8$ &$81.5$\\
%                     \hline
%         Ind-KNN       &  $84.3$ &  $83.1$ \\
%         Ind-KNN + hashing  &    $83.5$     &  $81.8$&\\
%         Ind-KNN + Reuse predictions& $85.0$     & $83.5$ \\
%         \hline
%     \end{tabular}
% }
% \label{tab: cifar10}
% \end{table*}

\begin{figure}[hbt]
\centering
\begin{subfigure}[h]{.47\linewidth}
    \centering\includegraphics[width=0.95\linewidth]{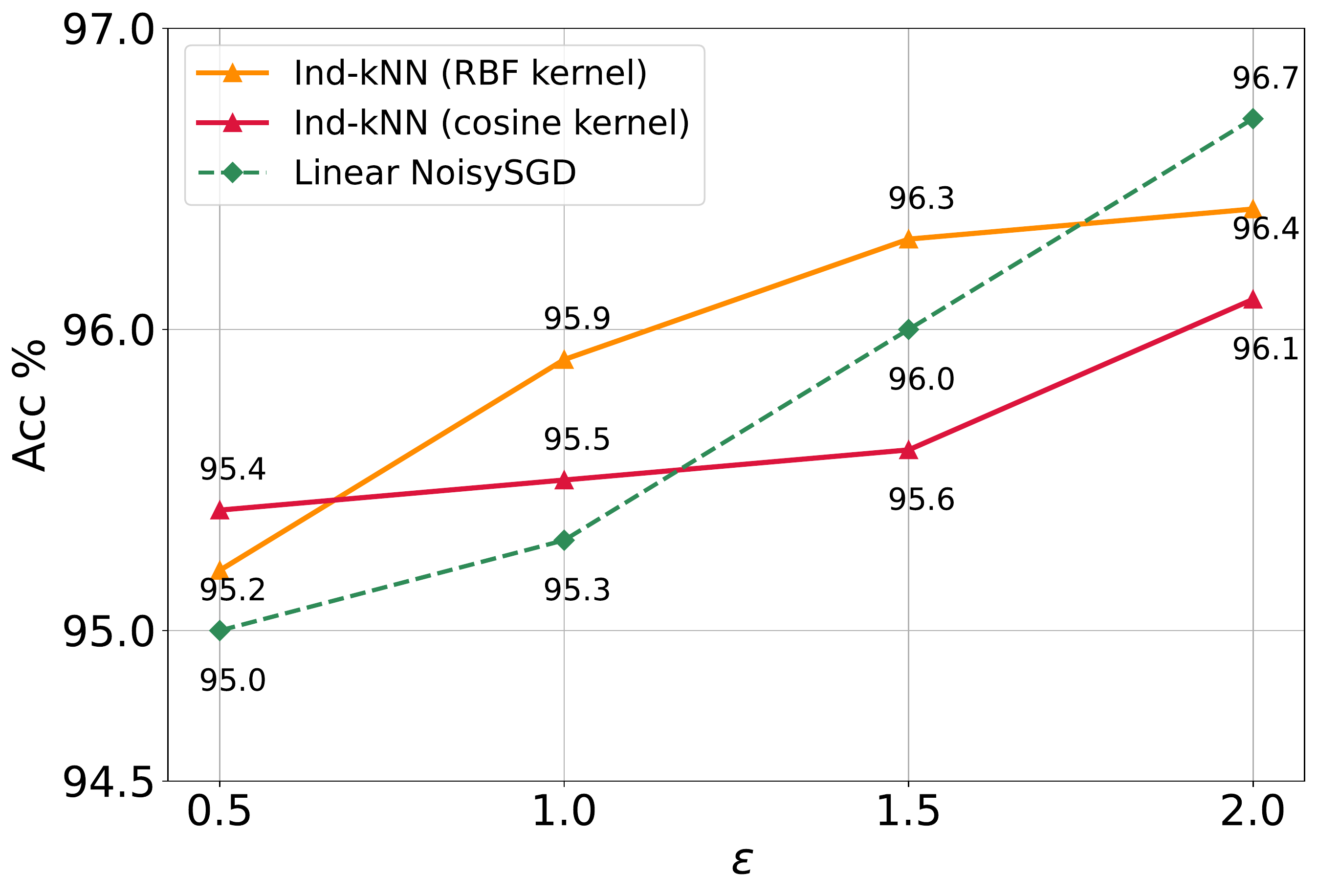}
    \caption{Privacy loss vs accuracy of $1000$ queries on CIFAR-10. }
    %\label{fig: cifar10_tradeoff}
\end{subfigure}
\begin{subfigure}[h]{.46\linewidth}
    \centering\includegraphics[width=0.95\linewidth]{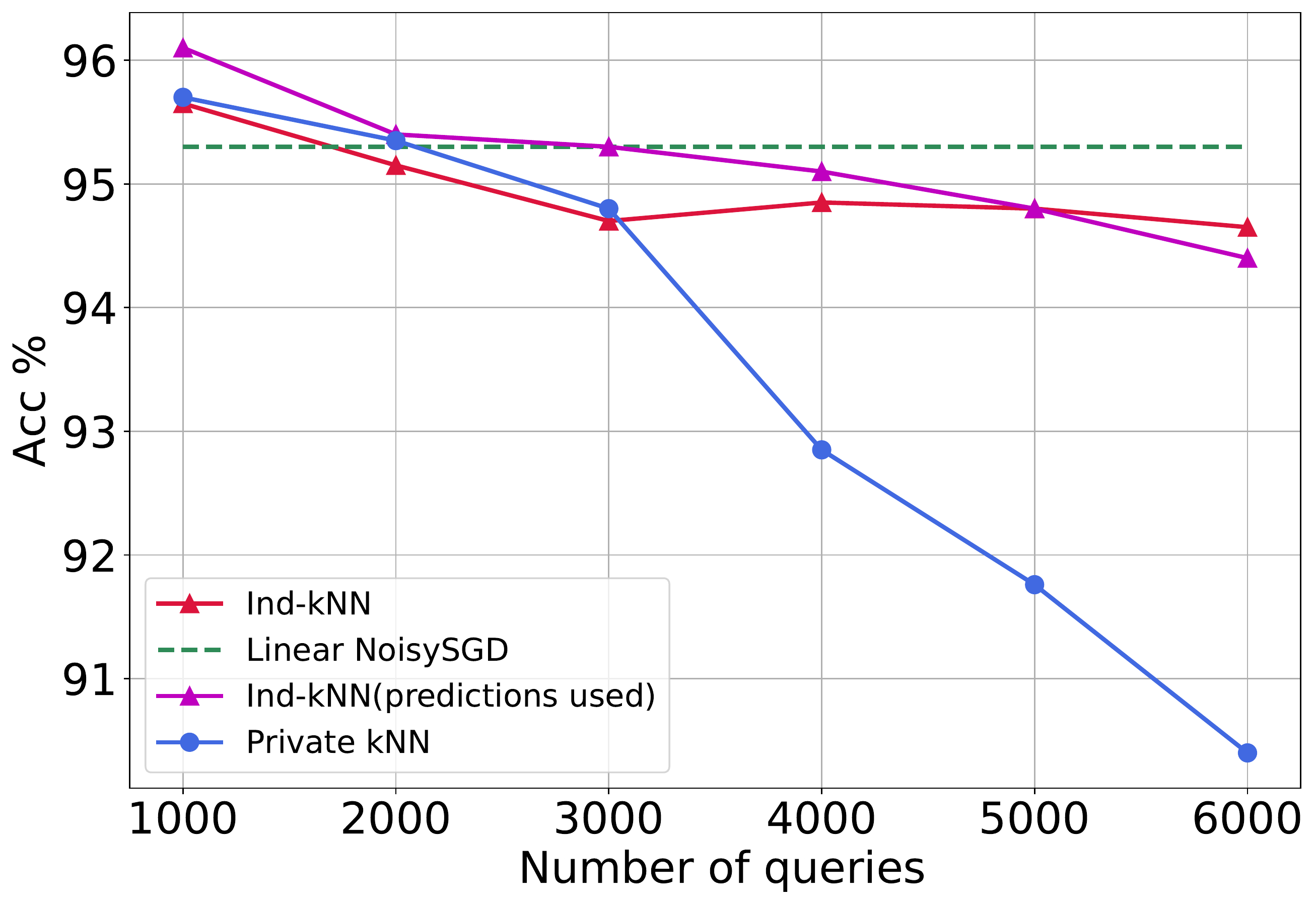}
    \caption{Accuracy vs number of query on CIFAR-10 under $(1, 10^{-5})$-DP. }
    \label{fig: query_vs_acc}
\end{subfigure}
\caption{Privacy-utility trade-offs on CIFAR-10. We plot the median accuracy across $5$ independent runs.}
\label{fig: cifar10} 
\end{figure}

% \paragraph{Hyper-parameters of Ind-KNN}
\noindent\textbf{Hyper-parameters of Ind-KNN.}
We set the individual RDP budget $B$ such that using RDP to DP conversion on $(\alpha, B\alpha)$-RDP satisfies the predefined privacy budget $(\epsilon, \delta)$. Then, we set the noise scale $\sigma_1$ to be $\sqrt{\frac{T}{6B}}$ to use roughly half of the individual RDP budget $B$ for each data point being selected at every query and tune the noise scale $\sigma_2$ on the validation set. We consider two kernel methods, the RBF kernel $\kappa(x, q) = e^{\frac{-||\phi(x)-\phi(q)||_2^2}{\nu^2}}$ and the cosine similarity $\kappa(x, q) = \cos(\phi(x), \phi(q))$. A linear scaling search is run on the minimum kernel weight threshold $\tau$ for each kernel method. Additional details are given in the  appendix.

%
%\begin{figure}
%    \centering
%    \includegraphics[width=0.99\linewidth]{ind_knn/img/cifar10_two.pdf}
 %   \vspace{-0.5em}
 %   \caption{
 %   Left: Privacy-accuracy trade-off of making $400$ predictions on CIFAR-10.  Right: Accuracy vs number of query on CIFAR-10 under $(2, 10^{-5})$-DP. We plot the median accuracy across $5$ independent runs.
 %   }
 %   \label{fig: cifar10}
 %    \vspace{-1em}
%\end{figure}

\begin{figure*}[t]
\centering	
\begin{subfigure}[t]{.3\linewidth}
    \centering\includegraphics[width=0.95\linewidth]{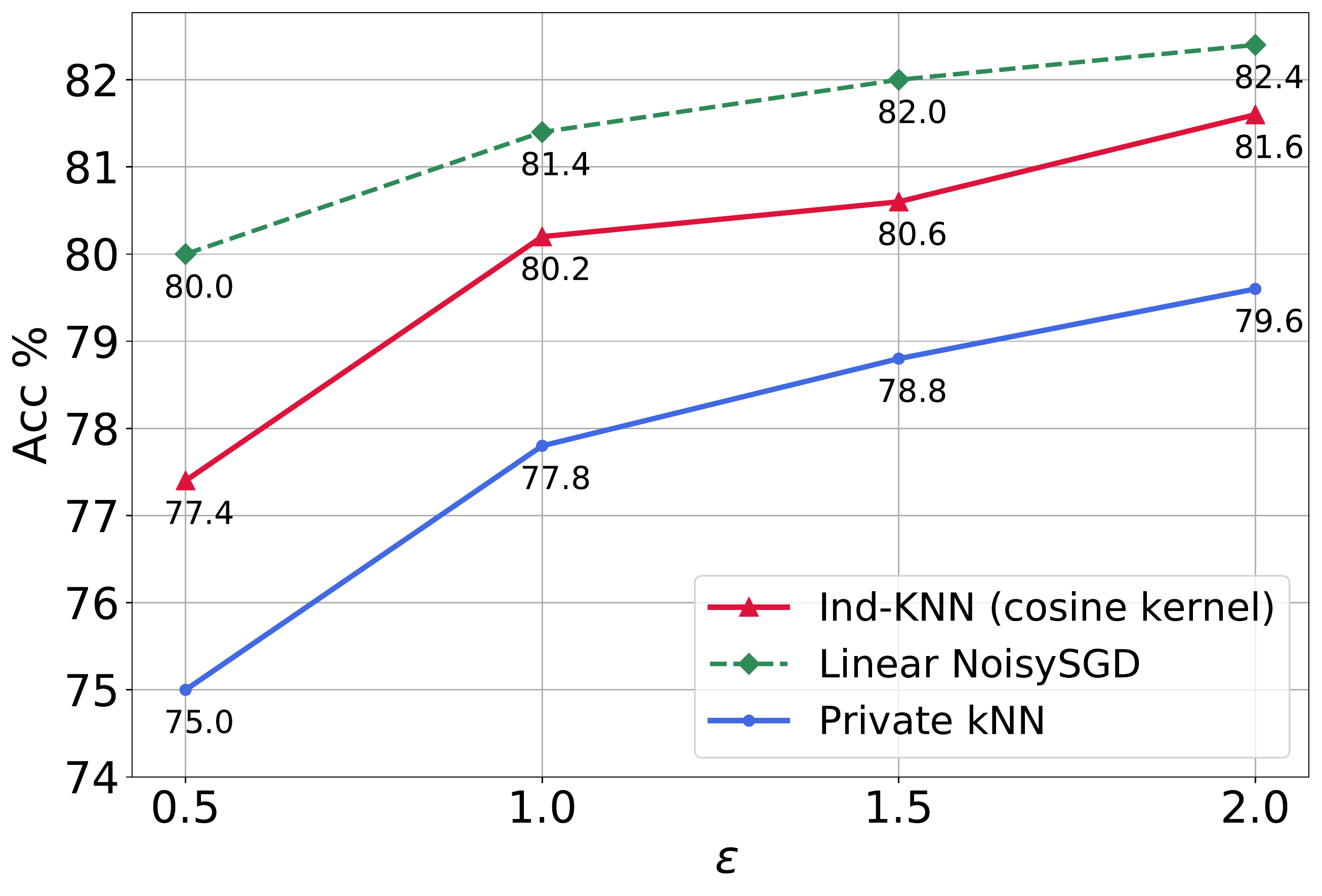}
    \caption{Accuracy of $500$ queries on FMNIST.}
    \label{fig:fmnist_tradeoff}
\end{subfigure}\qquad
\begin{subfigure}[t]{.3\linewidth}
    \centering\includegraphics[width=0.95\linewidth]{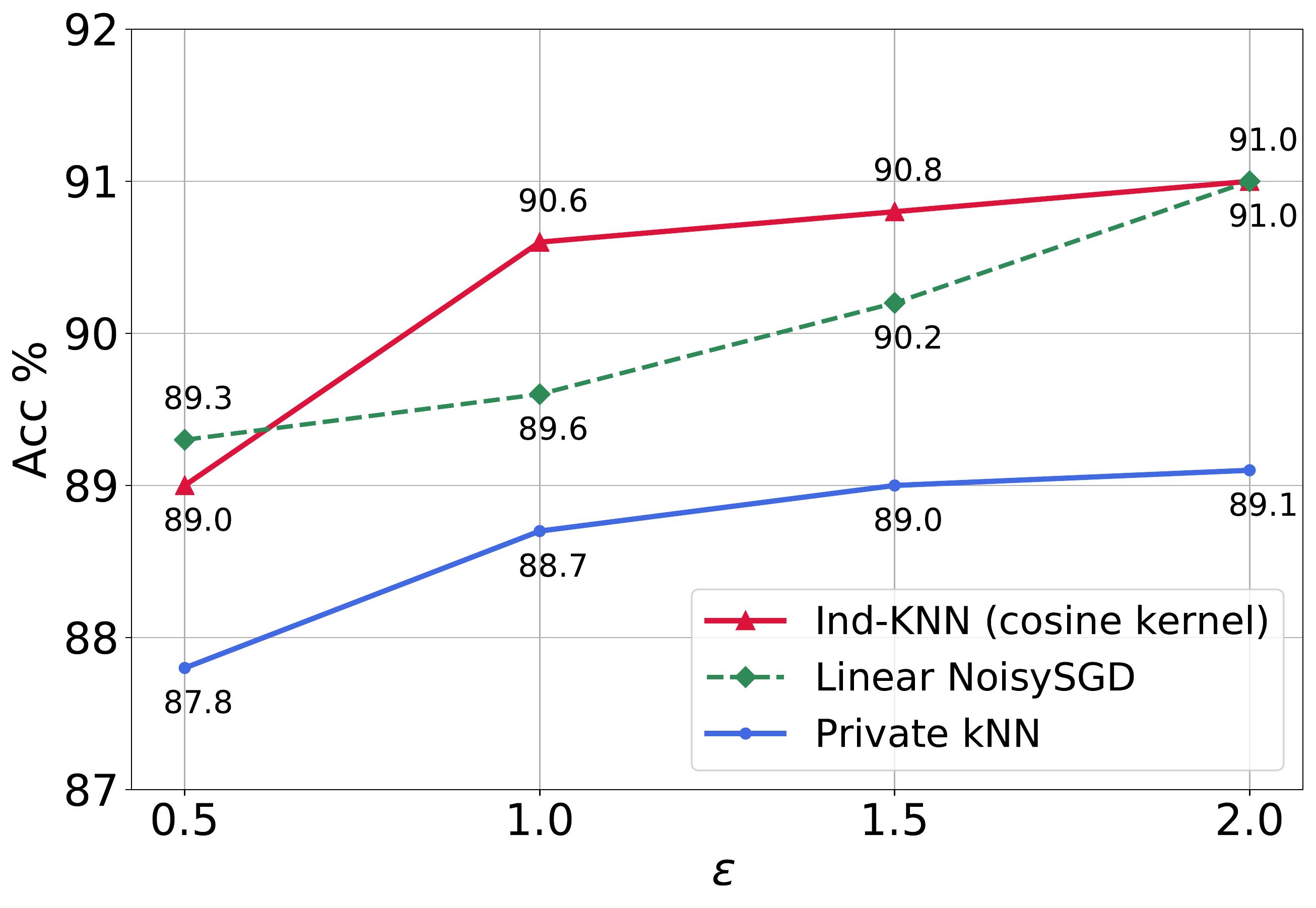}
    \caption{Accuracy of $800$ queries on AG News.}
    \label{fig: agnews_tradeoff}
\end{subfigure}\qquad
\begin{subfigure}[t]{.3\linewidth}
    \centering\includegraphics[width=0.95\linewidth]{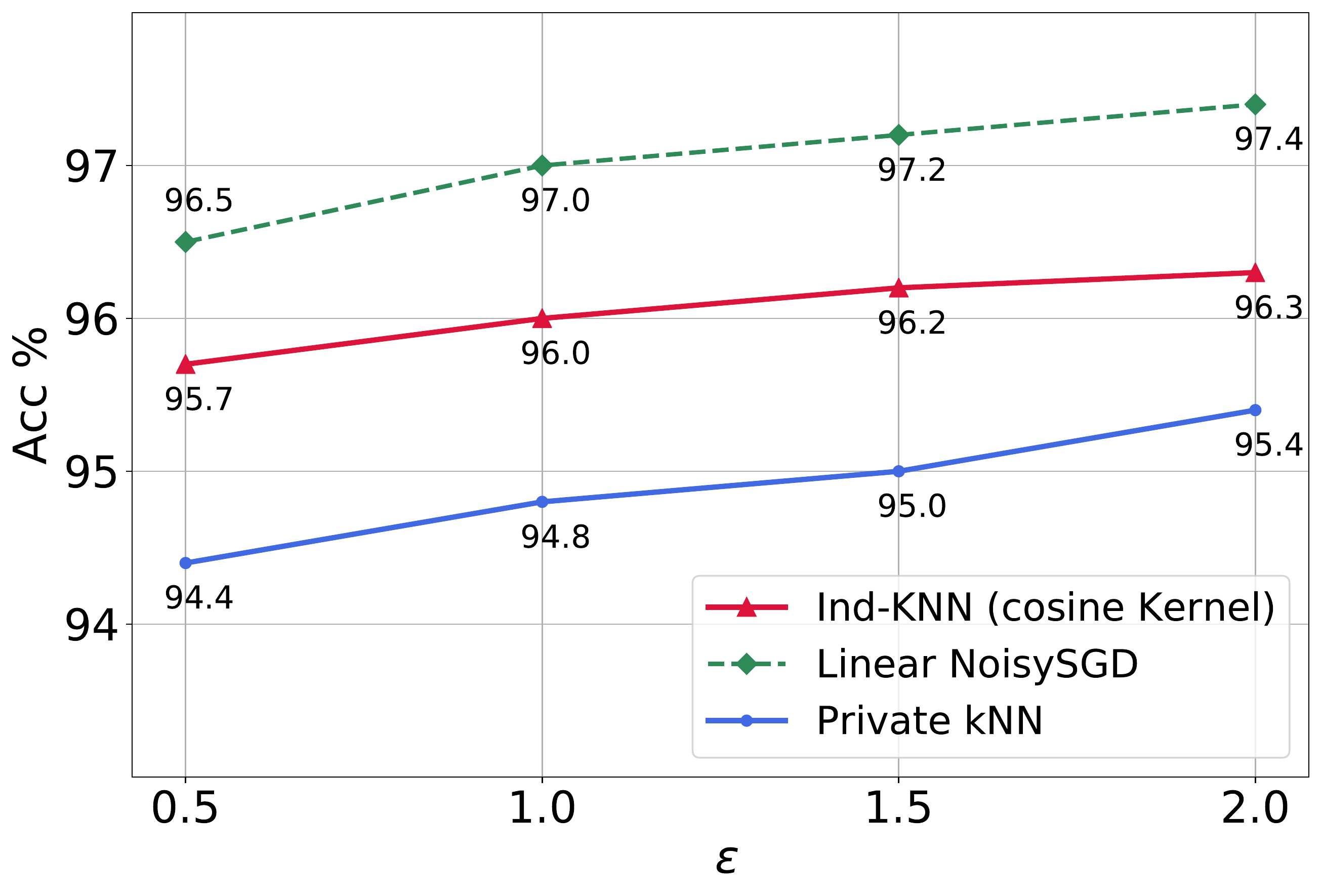}
    \caption{Accuracy of $800$ queries on DBPedia.}
    \label{fig: dbpedia_tradeoff}
\end{subfigure}
\caption{Privacy-accuracy trade-offs on FMNIST, AG News and DBPedia. We consider $\delta=10^{-5}$ for FMNIST and AG News and $\delta=10^{-6}$ for DBPedia.}
\label{fig: trade_off}

\end{figure*}
% \paragraph{Experiment setting}
\noindent\textbf{Experiment setting.}
For all experiments, we use a random seed to generate a validation set of size $T$. For example, we randomly sample $1000$ examples from the CIFAR-10 testing dataset and tune the best hyper-parameters of all approaches on the validation set. We then report the median accuracy across 5 independent sampled query sets. 
All experiments are conducted on a server with an Intel i7-5930K CPU @ 3.50GHz and Nvidia TITAN Xp GPU.

% \begin{figure*}[htb]
% \centering
% \begin{subfigure}[h]{.45\linewidth}
%     \centering\includegraphics[width=0.95\linewidth]{img/cifar10_hash.pdf}\label{fig: hash_cifar10}
%     \caption{Ind-KNN with LSH on CIFAR-10.}
% \end{subfigure}\qquad
% \begin{subfigure}[h]{.45\linewidth}
%     \centering\includegraphics[width=0.95\linewidth]{img/agnews_hash.pdf}\label{fig: hash_agnews}
%     \caption{Ind-KNN with LSH on Agnews.}
% \end{subfigure}
% \label{fig: hash}
% \end{figure*}

%\vspace{-0.5em}
\subsection{Main Results}\label{sec: exp_main_result}
% \vspace{-0.5em}
% \paragraph{Privacy-accuracy trade-off on CIFAR-10}
\noindent\textbf{Privacy-accuracy trade-off on CIFAR-10.}
In the top figure of Figure~\ref{fig: cifar10}, we plot the median accuracy evaluated on  $1000$ randomly chosen queries from the CIFAR-10 test set over a range of privacy budget $\epsilon$. The hyper-parameters were fine-tuned for each algorithm at each value of $\epsilon$.  For Ind-kNN, we found that the best hyper-parameter $\tau$ (the minimum threshold) increases as the privacy budget grows. We note this because, with smaller value of $\epsilon$, the added noise requires a larger margin among the selected neighbors' votes to  determine the correct output. This larger margin, in turn, corresponds to a smaller threshold and more selected neighbors.
For Ind-KNN with RBF kernel, we set the kernel bandwidth to $\nu=e^{1.5}$ and search for the optimal minimum threshold $\tau$ on the validation set. We find that different choices of kernel bandwidth in the RBF kernel produce similar accuracy results. 
As shown in Figure~\ref{fig: cifar10}, Ind-kNN with RBF kernel performs slightly better than its cosine kernel and both kernel methods are comparable to Linear NoisySGD across various value of $\epsilon$.

% For Private-kNN, we found that the best hyper-parameter $k$ decreases as the privacy budget increases. For example, we used $k=300$ when $\epsilon=0.5$, while $k=100$ when $\epsilon=2.0$.

% \paragraph{Accuracy vs $|$queries$|$ on CIFAR-10}
\noindent\textbf{Accuracy vs number of queries on CIFAR-10.}
Given a fixed privacy budget, the accuracy of all private prediction methods typically degrades as the number of predictions increases, while the accuracy of private training methods remains unaffected.   In the bottom figure of Figure~\ref{fig: cifar10}, we study how quickly the accuracy of Ind-KNN drops as the number of queries increases. We present the median accuracy of answering $T$ queries over five independent rounds. %We observe a low accuracy region for answering less than 400 queries, owing to the randomness associated with a small number of queries.
The accuracy of Private kNN drops rapidly with the increasing number of queries. This decline is expected, as Private kNN applies the standard R\'enyi composition theorem to analyze privacy loss, requiring the noise level to increase proportionally to the square root of $T$. In contrast, Ind-KNN uses individual privacy accountants, which only require selected neighbors to account for privacy loss, resulting in no significant accuracy drop as more queries are answered. Furthermore, exploiting released predictions allows Ind-KNN to answer an additional $1000$ queries (from $T=2000$ to $T=3000$) without an accuracy drop.
The figure also shows that if the number of queries is less than 2000, Ind-KNN can in fact outperform Linear NoisySGD, making it a practical alternative to private training methods when only a small number of predictions is needed.

% \paragraph{Privacy-accuracy trade-off on Fashion MNIST, AG News and DBPedia}
\noindent\textbf{Privacy-accuracy trade-off on Fashion MNIST, AG News and DBPedia.}
Next, we examine the privacy-accuracy trade-off on Fashion MNIST, AG News and DBPedia datasets. We use Ind-KNN with cosine kernel for all datasets. Figure~\ref{fig:fmnist_tradeoff} shows that Ind-KNN outperforms Private-kNN for all values of the privacy parameter $\epsilon$ on Fashion MNIST. On AG News, we compare the performance of Ind-KNN to that of Linear NoisySGD, and the results are presented in Figure~\ref{fig: agnews_tradeoff}. We evaluate $T=800$ queries on AG News and find that the accuracy of Ind-KNN either surpasses or matches that of Linear NoisySGD for $\epsilon\geq 0.5$. We also observe similar improvements over Private-kNN  on DBPedia.

Overall, Ind-KNN demonstrates its versatility by delivering competitive accuracy results on all three datasets, making it a promising solution for balancing differential privacy and accuracy.

 \begin{figure}[htb]
 \centering 
 \begin{subfigure}[c]{.46\linewidth}
     \centering\includegraphics[width=0.90\linewidth]{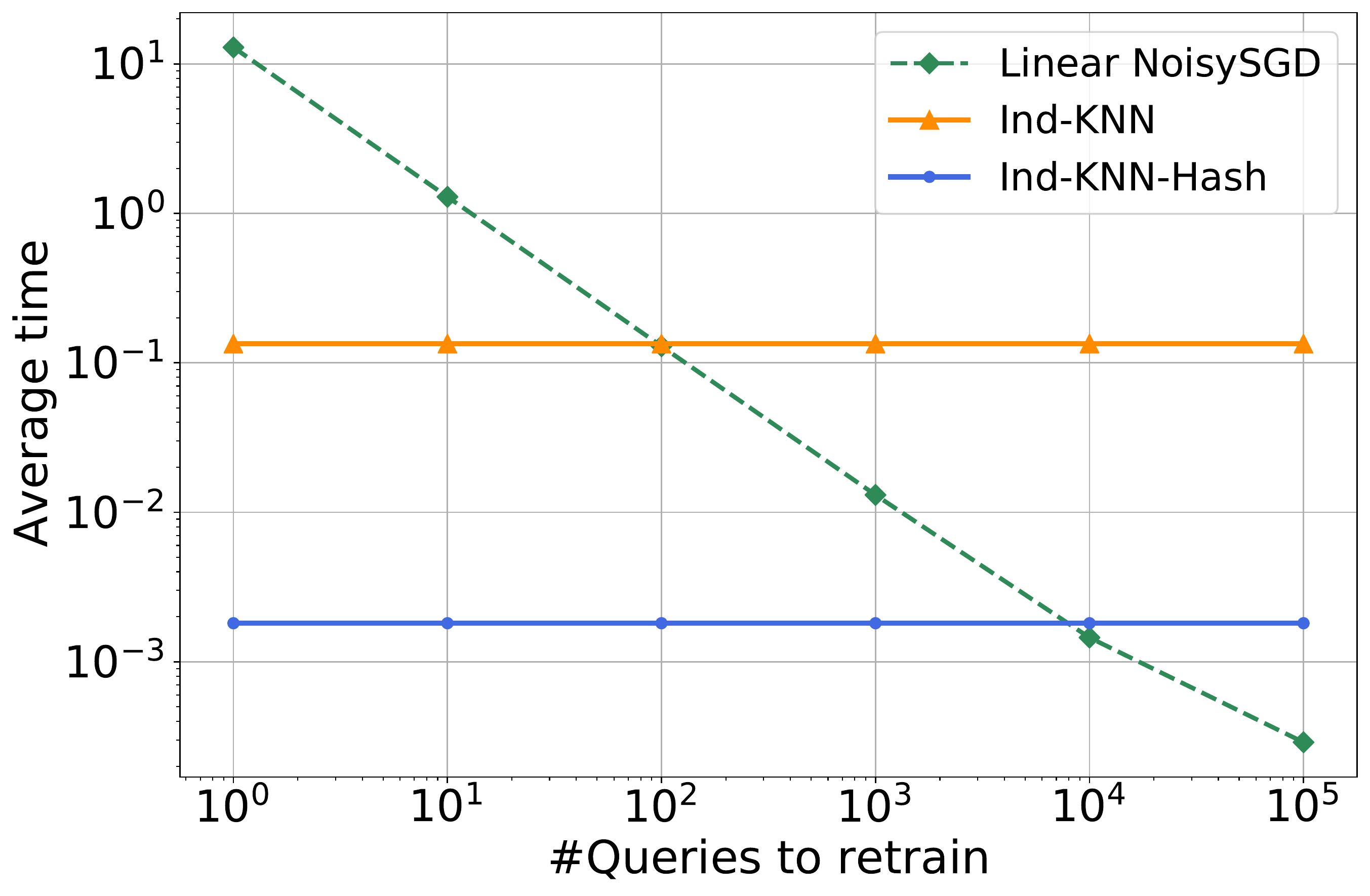}
    \caption{Amortized computational cost vs retraining frequency. \label{fig: speed}}
 \end{subfigure}\qquad
 \begin{subfigure}[c]{.46\linewidth}
     \centering\includegraphics[width=0.90\linewidth]{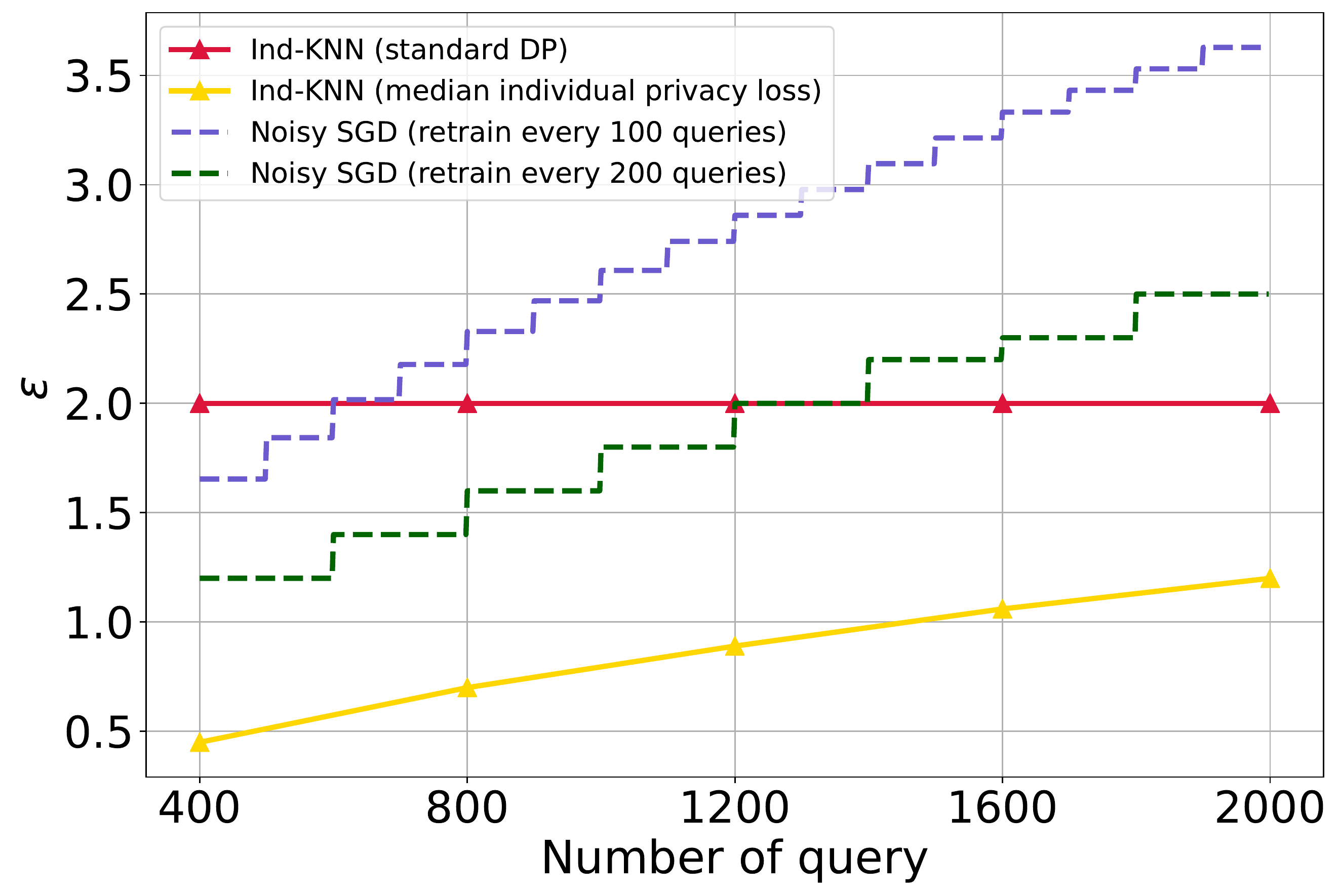}
    \caption{Privacy cost vs $|\text{queries}|$ when accuracy is aligned to $96.0\%$.}
     \label{fig: cost1}
      %\vspace{-0.5em}
 \end{subfigure}
 \caption{(a): We estimate the amortized computational cost by averaging the time (in seconds) spent to answer each query under different retrain settings on CIFAR-10. The x-axis denotes the retraining frequency, i.e., retraining a model every receiving $Q$ queries.  
(b): The accumulated privacy cost of answering a stream of $T=2000$ queries when the final accuracy (over 2000 queries) is aligned to $96.0\%$ on CIFAR-10. The red curve fixed the individual privacy budget at the beginning, resulting in a constant privacy loss. The yellow curve reports the median of individual privacy loss across  all private data. }\label{fig: unlearn}
\end{figure}

%\begin{figure}
 %   \centering
 %   \includegraphics[width=0.99\linewidth]{img/cifar10_unlearn.pdf}
  %  \caption{Left: The Averaged time spent (in second) to answer each query of different retrain settings on CIFAR-10. The x-axis denotes the retraining frequency, i.e., retraining a model every receiving $Q$ queries.  
   % Right: The accumulated privacy cost of answering a stream of $T=2000$ queries when the final accuracy (over 2000 queries) is aligned to $82.0\%$ on CIFAR-10. The red curve fixed the individual privacy budget at the beginning thus the privacy loss (converted from privacy budget) does not change. The yellow curve reports the median of individual privacy loss over all private data. 
 %   }
  %  \label{fig: unlearn}
%\end{figure}

\subsection{Ablation Studies}

We first perform an ablation study in Figure~\ref{fig: unlearn} to better understand how the  periodical retraining affects the performance of private training method and our Ind-KNN in terms of computational and privacy cost on CIFAR-10.
% \paragraph{Periodical retraining}

\noindent\textbf{Periodical retraining.}
In Figure~\ref{fig: speed}, we provide empirical measurements of the amortized computational cost associated with periodical retraining on CIFAR-10 of answering a stream of total $T=10^5$ queries. We assume a retraining request is triggered every time the model has answered $Q$ queries. To simplify the analysis, we assume each retraining is performed on the same dataset. For Linear NoisySGD, we retrain the model for 10 epochs and we calculate the per-query computational cost by dividing the total time spent on retraining and answering $T$ queries by $T$. This provides an estimate of the average time required to answer a single query.  For Ind-KNN with the cosine kernel, the average time of making predictions with is reported.   Ind-KNN-Hash uses 30 hash tables with the width parameter $b=8$. % The x-axis indicating the frequency of retraining requests $Q$.
 Our results demonstrate that the computational cost per query remains constant for Ind-KNN and Ind-KNN-Hash, as they do not require retraining the model, and the time required to add or delete individual data points is negligible. In contrast, for Linear NoisySGD, every retraining request incurs a substantial computational cost and the privacy loss grows $\propto \frac{1}{\sqrt{Q}}$ (proportional to the square root of total epochs). These findings highlight the advantage of Ind-KNN and Ind-KNN-Hash over Linear NoisySGD in terms of efficiency and resource utilization for machine unlearning and other scenarios with periodic retraining requests.

 \begin{figure}[htb]
 \centering 
 \begin{subfigure}[t]{.46\linewidth}
     \centering\includegraphics[width=0.90\linewidth]{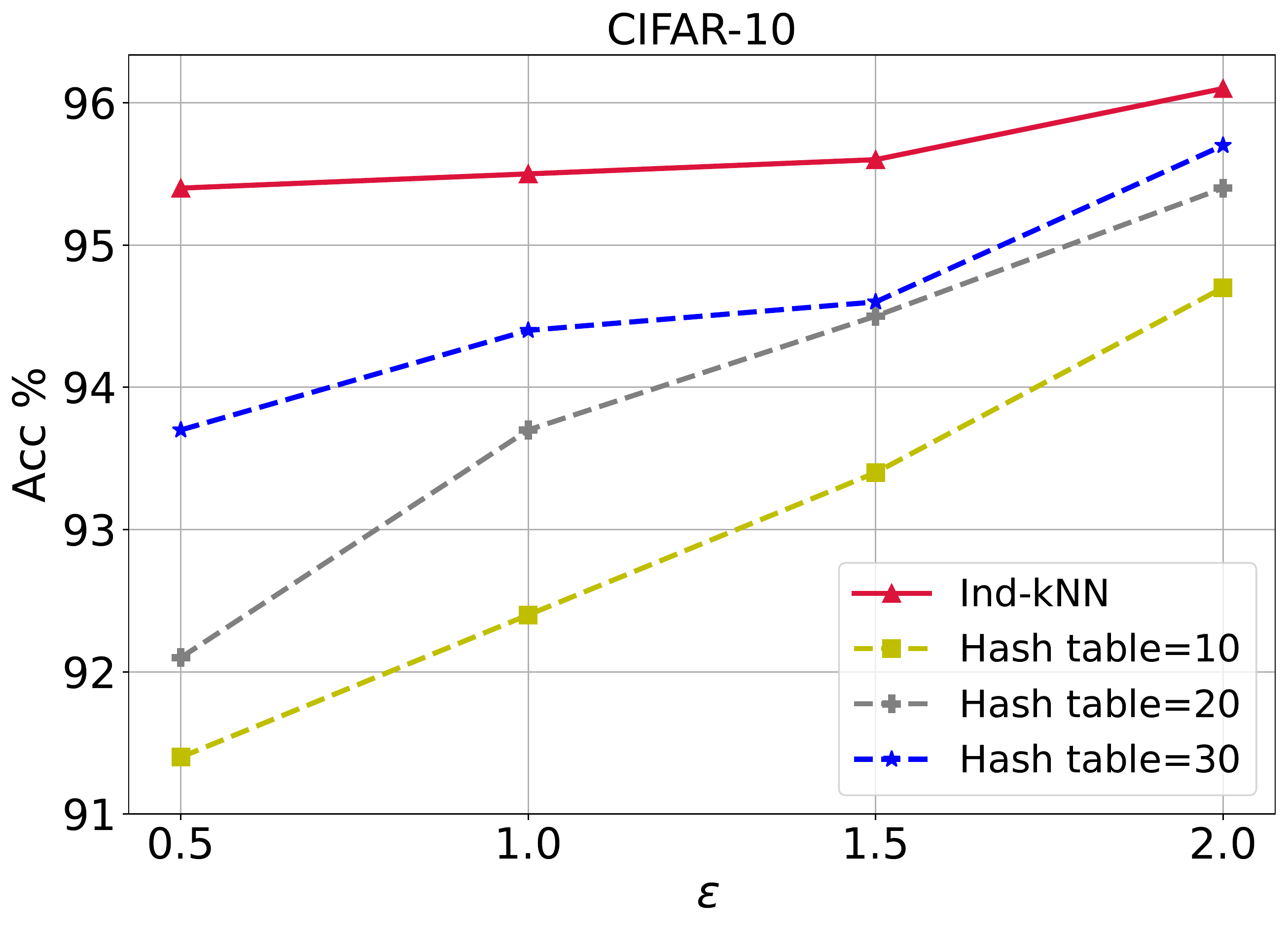}
    \caption{Accuracy of $T=1000$ queries on CIFAR-10.}
     \label{fig: hash_cifar}
 \end{subfigure}\qquad
 \begin{subfigure}[t]{.46\linewidth}
     \centering\includegraphics[width=0.90\linewidth]{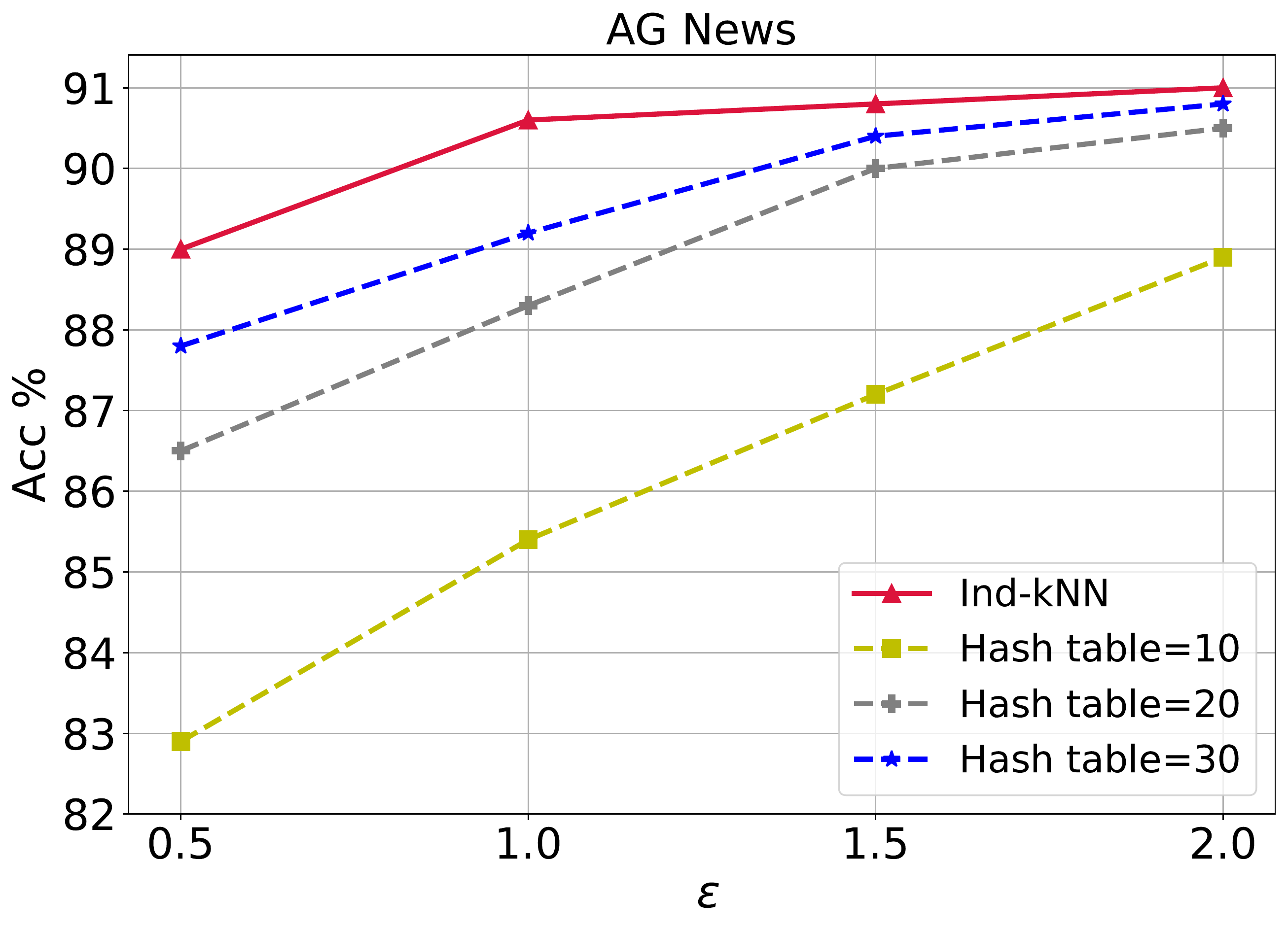}
    \caption{Accuracy of $T=1000$ queries on AG News.}
     \label{fig: hash_ag}
 \end{subfigure} %\vspace{-0.5em}
 \caption{Ablation study on hashing under $\delta=10^{-5}$. \label{fig: hash}}
 % \vspace{-1em}
\end{figure}
Figure~\ref{fig: cost1} evaluates the accumulated privacy loss of answering a stream of $T=2000$ queries on CIFAR-10.
We tune hyper-parameters for both approaches such that the averaged accuracy of answering $T$ queries is aligned to $96.0\%$. We consider two types of retraining scenarios: $Q=100$ and $Q=200$. 
Periodic retraining has a negligible privacy impact on  Ind-KNN. Therefore, we only use one red curve to indicate the privacy loss of Ind-KNN under two scenarios. The individual privacy budget of Ind-KNN is pre-determined, thus the standard privacy guarantee remained unchanged when making more predictions. The yellow curve plots the median of individual privacy loss over all private data points and reflects how much individual privacy loss deteriorates as the number of answered queries increases. We note the median individual privacy loss is $\epsilon=1.2$ after answering 2000 queries, which suggests that only half of the privacy budget has been spent at an individual level.
The privacy loss curve of Ind-KNN and two Linear NoisySGDs are met when there received six retraining requests. This suggests that if there are more than six retraining requests among the $2000$ queries, the privacy loss of Ind-KNN would be  better than that of Linear NoisySGD. 
% what is yellow curve?

\begin{table}[ht]
\centering
\caption{ Test Accuracy of $T=1000$ queries on CIFAR-10 under different pre-trained models: vision transformer (ViT)~\citep{vit},  SimCLRv2 model~\citep{clr} and ResNet50~\citep{He_2016_CVPR}. }
\begin{tabularx}{\linewidth}{C|C|C|C|C}
  \hline
   $\epsilon (\delta = 10^{-5})$ & Method & ResNet50 & SimCLRv2 & ViT \\
  \hline
  \multirow{3}{*}{$\epsilon=0.5$} & Linear NoisySGD  & 86.2\% & 89.7\%& 95.0\%  \\
    & Private kNN &73.1\%& 76.0\% &  94.4\% \\
      & Ind-kNN  &79.4\%& 82.4\%& 95.2\%  \\
  \hline
    \multirow{2}{*}{$\epsilon=2.0$} & Linear NoisySGD &88.4\% & $90.2\%$ & $96.7\%$  \\
    & Private kNN &81.6\% & 84.7\%& $96.3\%$ \\
      & Ind-kNN  & 82.8\% & 86.3\%& $96.4\%$  \\
       \hline
        \multirow{2}{*}{$\epsilon=\inf$} & Linear NoisySGD & 90.0\% & 90.7\% & 97.0\% \\
    & Private kNN & 82.9\% &  85.1\% & 96.6\%\\
    & Ind-kNN  &  84.7\% & 89.2\% &  96.9\% \\
    \hline
\end{tabularx}%
\label{tab: vary_model}
\end{table}

\begin{table*}[ht]
\centering
\caption{The Averaged time (in second) to answer each query on CIFAR-10 and AG News using Ind-KNN and its hashing variants.}
\begin{tabularx}{\linewidth}{l|C|C|C|C}
    \toprule
    Dataset & Table=10& Table=20 &Table=30 &Ind-KNN  \\
    \hline
      CIFAR-10    &   0.02 & 0.03&  0.04& 0.25 \\ 
      \hline
     AG News &0.01 & 0.02 & 0.03 & 0.29\\
    \hline
\end{tabularx}

\label{tab: time_vs_hash}
\end{table*}

% \paragraph{Computational cost vs utility}
\noindent\textbf{Ablation study on hashing.}
% what's computational cost?
In Sec~\ref{sec: hash}, we introduce hashing to improve the computational efficiency of Ind-KNN. We now investigate the trade-off between computational cost and utility of Ind-KNN-Hash on CIFAR-10 and AG News.  We set the width parameter $b=8$ for CIFAR-10 and $b=9$ for AG News, and evaluate the  performance of Ind-KNN-Hash with varying number of hash tables.
As shown in Figure~\ref{fig: hash} and Table~\ref{tab: time_vs_hash}, the accuracy of hashing variants increases with more hash tables and more computational cost. We note that the computational cost roughly grows linearly with the number of the hash table.
In particular, Ind-KNN with $30$ hash tables matches the accuracy of the original Ind-KNN for a wide range of epsilon on CIFAR-10 but reduces the running time per query from $0.25$ second to $0.03$ second. 
Figure~\ref{fig: agnews_tradeoff} shows similar observations for AG News.

%\begin{figure}[ht]
%\centering
%\includegraphics[width=0.99\linewidth]{img/res_hash.pdf}
%\caption{Ind-KNN with LSH on CIFAR-10 and AG News.}
%\label{fig: hash_agnews_cifar10}
%\end{figure}

% what's the impact of changing the dataset size?

\noindent\textbf{Ablation study on the feature extractor.}
The quality of the feature extractor plays an crucial role in all three pre-trained feature-based methods. Remarkably, With the ViT feature extractor, even the Private kNN achieves an impressive accuracy of $96.3\%$ at $\epsilon=2.0$ on CIFAR-10,  surpassing the previously reported best result of $95.4\%$~\citep{de2022unlocking} achieved using Wide-ResNets. Next, we present an ablation study focusing on three feature extractors and investigate the efficiency of each method on the CIFAR-10 task. Specifically, we consider three widely used vision models:  vision transformer (ViT) ~\citep{vit}, the SimCLRv2 model~\citep{clr} and ResNet 50~\citep{He_2016_CVPR}. The SimCLRv2-based feature extractor has been considered by prior work Linear NoisySGD (\cite{tramer2020differentially}), which trains a ResNet model on unlabeled ImageNet using SimCLRv2 model and provides a $4096$-dim feature for each input image. For Resnet50, we consider the publicly-available ImageNet-pretrained Reset50 from Pytorch, which achieves a non-private accuracy at $90.0\%$ for LinearSGD.  As shown in Table~\ref{tab: vary_model}, we find that Linear NoisySGD  outperforms Private kNN and our Ind-kNN across ResNet50 and SimCLRv2. However, the performance gap decreases when applying a better feature extractor.  This can be explained by the fact of their non-private performance. We also note that Private kNN is more fragile when $\epsilon$ is small, which could be due to its ``loose'' privacy analysis.   Meanwhile, Ind-kNN handle the setting of small $\epsilon$ nicely, and can sometimes outperform Linear NoisySGD with a good feature extractor. 

%Additionally, the choice of the width parameter also has an impact on the performance, e.g., a smaller width value generally leads to higher accuracy but increased computational cost. The ablation study on the width parameter is deferred to the appendix.

%% file: 09_summary.tex
%\vspace{-1.em}
\section{Summary}
 %\vspace{-1.em}
The paper proposes a new algorithm, Individual Kernelized Nearest Neighbor (Ind-KNN), for private prediction in machine learning that is more flexible and updatable over dataset changes than private training. By modifying the KNN prediction and leveraging individualized privacy accountants, Ind-KNN allows a precise control of privacy at an individual level. Through extensive experimentation on four datasets, we demonstrate that Ind-KNN outperforms prior work Private kNN in terms of privacy and utility trade-offs. Furthermore, Ind-KNN exhibits superior computational efficiency and utility when dealing with frequent data updates, surpassing the private training method.

%% file: 010_appendix.tex
\section{Omitted proofs and algorithm in Section 3}
\begin{theorem}[Restatement of Theorem 3.1]
Algorithm 3 satisfies $(\alpha, B\alpha)$-RDP for all $\alpha\geq 1$.
\end{theorem}

\begin{proof}

The privacy analysis relies on  individual RDP (Definition 2.4), which quantifies the maximum impact of adding or deleting a specific individual from any potential dataset to the prediction outcome, measured in terms of R\'enyi divergence.

We first demonstrate that only the selected neighbors have to account for their individual privacy loss. 
The decision rule for ``being selected'' is
based on a comparison between the kernel weight and a data-independent threshold $\tau$, which is not influenced by any other private data points. Therefore, ``unselected'' neighbors do not incur any individual privacy loss. 

For each selected neighbor $(x_i, y_i)$ at time $t$, its individual privacy analysis is broken down into two parts: the first part is the release of the number of neighbors $|\cN_t|$, and the second part is the release of its label associated with the kernel weight.

Note that adding or removing one selected neighbor would only change $|\cN_t|$ by $1$, thus the individual RDP of releasing $|\cN_t|$ at order $\alpha$ satisfies $\frac{\alpha}{2\sigma_1^2}$-RDP for all selected data. 
We next analyze the individual RDP of releasing the label.
Fix a selected neighbor $z=(x_i, y_i)$, for all possible set of selected neighbors $\cN_t=(z_1, ..., z_m)$ that include $z$, it holds that
\begin{align*}
D_\alpha^{\leftrightarrow}\bigg(\big(\sum_{j\in \cN_t} \kappa(x_j, q_t)\cdot y_j\big)+\cN(0, \sigma_2^2K_t \mathbb{I}_c)||\big(\sum_{j \in \cN_t \setminus z} \kappa(x_j, q_t)\cdot y_j\big)+\cN(0, \sigma_2^2K_t \mathbb{I}_c) \bigg)\leq \frac{\kappa(x_i, q_t)^2\alpha }{2\sigma_2^2 K_t}
\end{align*}
by the definition of individual RDP.

Finally, The ``delete'' step in the algorithm ensures that the privacy loss for each private data point $(x_i, y_i)$ is bounded by a fixed value $B$, i.e., $\sum_{j=1}^t \left((g_i+\frac{1}{2\sigma_1^2\cdot K_j})\cdot \mathbb{I}[(x_i, y_i) \in \cN_j]\right)\leq B$. 
According to the fully adaptive composition theorem of individual RDP (Theorem 2.5),
by ensuring the  sum is less than or equal to $B$ for all time steps $t$ and for all data point $(x_i, y_i)$, the algorithm is shown to be $(\alpha, \alpha\cdot B)$-RDP.
\end{proof}

We show the full algorithm of Ind-KNN-Hash in Algorithm \ref{alg:ind_kNN_hashing}.

\begin{algorithm}[t]
\caption{Ind-KNN-Hash}
\label{alg:ind_kNN_hashing}
\begin{algorithmic}[1]
\STATE{\textbf{Input}: Dataset $S\in (\cX \times \cY)^n$, number of hash tables $L$ and the width parameter $b$,
the kernel function $\kappa(\cdot, \cdot)$, the minimum kernel weight threshold $\tau$, sequence of queries $q_1, ..., q_T$, the noisy scale $\sigma_1$ and $\sigma_2$ and the individual budget $B$. }
\STATE{Initialize individual budget $z_i = B, \forall i \in [n]$.}
\STATE{Construct a LSH family: $\cF =(f_1, ..., f_L)$, where $f_\ell: \cR^d \to \{0, 1\}^b$.}
\FOR{$t=1$ to $T$}
\STATE{Retrieve the hash set: $\cF(q_t)$.}
\STATE{Update the active set $S=\{(x_i, y_i)|z_i >0, (x_i, y_i) \in \cF(q_t)\}$.}
% \STATE{$a_t$ = Algorithm~\ref{alg:noisy_label}($S, \kappa(\cdot, \cdot), q_t, \sigma_1, \sigma_2$).}
\STATE{The selected neighbors: $\cN_t := \{(x_i, y_i)|\kappa(x_i, q_t) \geq \tau   \text{ for all } i \in S\}$.}
\STATE{Drop $(x_i, y_i)$ from $\cN_t$ if $z_i \leq \frac{1}{2\sigma_1^2}$.}
\STATE{Release $|\cN_t|$: $K_t:=|\cN_t| + \cN(0, \sigma_1^2).$}
\FOR{$(x_i, y_i)\in \cN_t$}
\STATE{Update $z_i$ after releasing $K_t$: $z_i=z_i - \frac{1}{2\sigma_1^2}$. }
\STATE{Evaluate individual ``contribution'': $g_i = \min \left(\frac{\kappa(x_i, q_t)^2}{2\sigma_2^2\cdot K_t}, \sigma_2\sqrt{2 K_t z_i}\right)$.}
\STATE{Update $z_i$ after releasing label: $z_i = z_i - g_i$.}
%\STATE{Drop $(x_i, y_i)$ from the active set $S$ if $z_i\leq0$.}
\ENDFOR
\STATE{Compute $a_t = \argmax_{j \in [c]}\big(\sum_{i\in \cN_t}\kappa(x_i, q_t)\cdot y_i +\cN(\boldsymbol{0}, \sigma_2^2\cdot K_t\mathbf{1}_c)\big)_j$.}
\ENDFOR
\STATE{\textbf{Return} $(a_1, ..., a_T)$}
\end{algorithmic}
\end{algorithm}

\begin{figure*}[h]
 \centering 
\begin{minipage}{0.45\textwidth}
     \centering\includegraphics[width=0.98\linewidth]{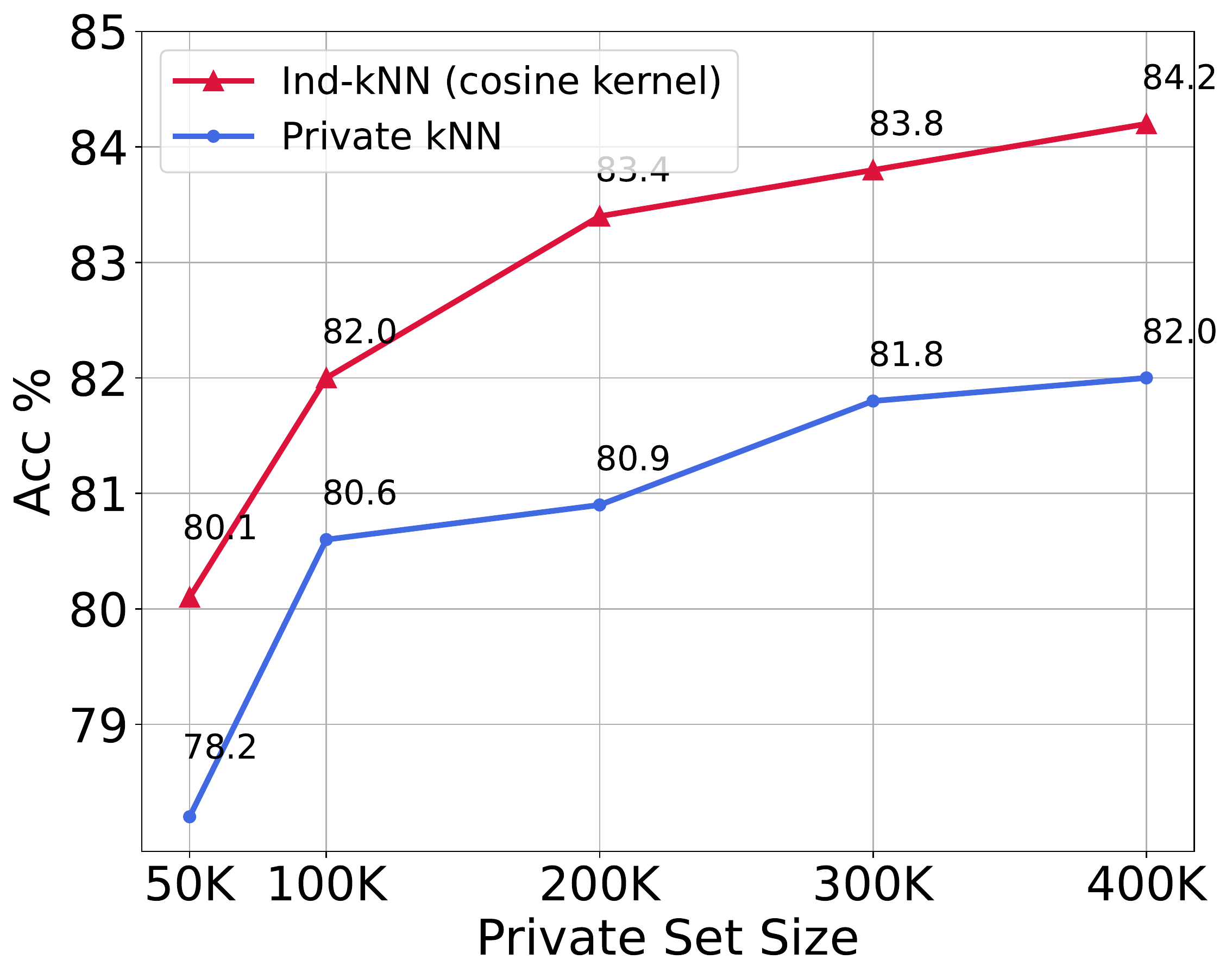}
    \caption{Accuracy of answering $T=2000$ queries for a private dataset of size $N$ under $(\epsilon=1., \delta= 1/N)$-DP. We use the ResNet50 pre-trained model as the feature extractor. }
     \label{fig: repeat}
\end{minipage}
% \vspace{-1em}
\end{figure*}

\section{More experiments}

\noindent\textbf{Ablation study on the size of private data set  }
We present an ablation study on the size of private dataset, where we simulate private datasets of varying sizes $N \in \{50K, 100K, 200K, 300K, 400K\}$ by replicating the training set of the CIFAR-10 dataset. Figure~\ref{fig: repeat} plots the accuracy of answering $T=2000$ queries with each private dataset size under $(\epsilon=1.0, \delta = \frac{1}{N})$-DP with the Resnet50 feature extractor. 
Both prediction approaches can benefit from an increase in private data: Private kNN can take advantage of subsampling while Ind-KNN could leverage a larger active set. The figure demonstrates that our Ind-KNN scales well with more private data, enabling it to address even millions of queries when the private dataset is billion-scale.

\textbf{Ablation study on the threshold $\tau$.}
The minimum kernel weight threshold $\tau$ determines the number neighbors selected for each query-response pair. We conduct an ablation study to investigate the relation between the optimal $\tau$ and the privacy level. Table~\ref{tab: best_hyper_indknn} provides the set of hyper-parameters of Ind-kNN that results in the best utility. Our finding shows that the optimal choice on $\tau$ increases as $\epsilon$ increases across four datasets and two kernel methods. We conjecture this is because when  $\epsilon$ is small, the added noise requires a larger margin among the top-k votes to determine the correct output, thus requiring a smaller $\tau$. In contrast, when $\epsilon$ is large, the smaller noise scale enables the algorithm to  pick a set of more selective neighbors, thus resulting in a larger $\tau$.

%In Table~\ref{tab: cifar10}, we provide an ablation study of two variants of Ind-KNN under $(2.0, 10^{-5})$-DP. We instantiate all Ind-KNN variants with cosine kernels, and the hashing variant uses $12$ hash tables with the width parameter $b=8$. When answering $400$ queries, we find that using hashing decreases the accuracy from $84.3\%$ to $83.5\%$, which still matches the accuracy of $83.4\%$ from Linear NoisySGD. Moreover, exploiting released predictions enables Ind-KNN to have better accuracy when answering both $400$ and $800$ queries. 
\begin{comment}
\begin{table}[ht]
\centering
\setlength{\tabcolsep}{3pt}
\caption{Median accuracy on CIFAR-10 under $(2.0, 10^{-5})$-DP.  We report the accuracy (\%) of 400 and 800 test queries.}
\begin{tabular}{lcccc}
    \toprule
    Methods & \  Acc ($400$) & Acc ($800$)  \\
    \hline
     Linear NoisySGD      & $83.4$  & $83.4$  \\ 
     Private kNN &$82.8$ &$81.5$\\
    \hline
    Ind-KNN       &  $84.3$ &  $83.1$ \\
    Ind-KNN-Hash  &    $83.5$     &  $81.8$&\\
    Ind-KNN + Reuse predictions& $85.0$     & $83.5$ \\
    \hline
\end{tabular}
\label{tab: cifar10}
\end{table}
\end{comment}
\begin{table}[ht]
\centering
\caption{The range of hyper-parameters for Private kNN.}
\begin{tabular}{c|cccc}
    \toprule
   Hyper-parameters & CIFAR-10& Fashion MNIST & AG News & DBpedia \\
    \hline
 sampling ratio $p$ & \multicolumn{4}{c}{$\{0.02, 0.05, 0.1, 0.2\}$}\\
      \hline
 number of neighbors $K$ & $\{100, 200, 300, 400\}$ & $\{100, ..., 500\}$ & $\{100, ..., 600\}$& $\{100, ..., 600\}$\\
    \hline
\end{tabular}
\label{tab: hyper_private_knn}
\end{table}
\section{Experimental details}
In this section, we present the implementation details of Ind-KNN and Private kNN.

\textbf{Hyper-parameters search of Ind-kNN.}
%We set the individual RDP budget $B$ such that using RDP to DP conversion on $(\alpha, B\alpha)$-RDP satisfies the predefined privacy budget $(\epsilon, \delta)$. 
The noise scale $\sigma_1$ is set to be $\sqrt{\frac{T}{6B}}$ to use roughly half of the individual RDP budget $B$ for each data point being selected at every query. Further reducing $\sigma_1$ does not result in significant improvement. To prevent overflow in $\frac{\kappa(x_i, q_t)^2}{2\sigma_2^2K_t}$ due to random noise, $K_t$ is set to $\max(K_t,30)$ for all experiments. 
For Ind-KNN using a cosine kernel, we fine-tune the noise scale $\sigma_2$ and the threshold $\tau$ on the validation set. To reduce the computational cost of searching all possible $(\sigma_2, \tau)$-pairs, we first estimate the optimal threshold $\tau$ by running a  non-private Ind-KNN on the valid set to collect individual kernels weights and sweep through $\tau \in \{0.05, 0.1, ..., 0.95\}$. With $\tau^*$ in hand, we perform a second round of hyper-parameter search for the optimal $(\sigma_2, \tau)$ pair under different privacy levels, where $\tau$ ranges between $[\tau^*-0.05, \tau^*+0.05]$. The table below records the range of $(\sigma_2,\tau)$ pairs consider in the second-round  search.

We present the range of hyper-parameters search for  Private kNN in Table~\ref{tab: hyper_private_knn} and the best hyper-parameter sets we use in Table~\ref{tab: best_hyper_indknn}.

\textbf{Feature preprocessing.}
As we mentioned in the experiment section, we use a pre-trained ResNet50 model on ImageNet for feature extraction in image classification tasks. Then we perform L2 normalization on the extracted features as a preprocessing step. As for text classification, the extracted features from sentence-transformer are already normalized. In this case, we don't need to apply any additional preprocessing steps.
% how to pre-process features? 
% the hyper-parameter ranges for all mechanisms.
% the best hyper-parameters.
% ablation study on tau

\begin{table}[ht]
\centering
\caption{Set of hyper-parameters of Ind-KNN resulting the best utility for a set of privacy budgets used in Sec 4.2. We apply the ViT-based feature extractor for CIFAR-10, the pre-trained ResNet50 for Fashion MNIST and the sentence-transformer for AG News and DBpedia. }
\resizebox{1.0\columnwidth}{!}{
\begin{tabular}{c|c|cccc}
    \toprule
 Methods & Datasets & $\epsilon=0.5$& $\epsilon=1.0$ & $\epsilon=1.5$  & $\epsilon=2.0$\\
    \hline
\multirow{3}{*}{Cosine kernel ($\sigma_2, \tau$) }& CIFAR-10 & $(\sigma_2=0.7,\tau=0.12)$ &$(\sigma_2=0.4, \tau=0.12)$ &$(\sigma_2=0.4, \tau=0.12)$ &$(\sigma_2=0.4, \tau=0.12)$\\
& Fashion MNIST&($\sigma_2=1.3, \tau=0.6$) &$(\sigma_2=0.6,\tau=0.6)$ & ($\sigma_2=0.3, \tau=0.6$)& $(\sigma_2=0.3, \tau=0.6)$\\
& AG News& ($\sigma_2=0.6, \tau=0.35$) &($\sigma_2=0.4, \tau=0.36$) & $(\sigma_2=0.25, \tau=0.37)$&$(\sigma_2=0.2, \tau=0.38)$\\
& DBpedia& ($\sigma_2=0.45, \tau=0.35$) & $(\sigma_2=0.3,\tau=0.37)$ &$(\sigma_2=0.2, \tau=0.37)$ & $(\sigma_2=0.1, \tau=0.38)$ \\
    \hline
    \multirow{2}{*}{RBF kernel $(\sigma_2, \tau, \nu)$}& CIFAR-10& $(\sigma_2=0.6, \tau=0.8)$ &$(\sigma_2=0.5, \tau=0.25)$ &$(\sigma_2=0.4, \tau=0.26)$ &$(\sigma_2=0.2, \tau0.28)$\\
& Fashion MNIST & ($\sigma_2=1.3,\tau=0.83$)&$(\sigma_2=0.7,\tau=0.82)$ & $(\sigma_2=0.4, \tau=0.84)$& ($\sigma_2=0.3, \tau=0.84$)\\
    \hline
\multirow{2}{*}{ Hash $(L=30, b, \sigma_2, \tau)$}& CIFAR-10 & $(b=8, \sigma_2=0.6, \tau=0.25)$ &$(b=8, \sigma_2=0.4, \tau=0.50)$ &$(b=8, \sigma_2=0.3, \tau=0.52)$ &$(b=8, \sigma_2=0.2, \tau=0.53)$\\
& AG News & ($b=9, \sigma_2=0.7, \tau=0.35$)& ($b=9, \sigma_2=0.4, \tau=0.36$) & ($b=9, \sigma_2=0.25, \tau=0.36$) &($b=9, \sigma_2=0.2, \tau=0.36$)\\
    \hline
\end{tabular}}
\label{tab: best_hyper_indknn}
\end{table}

\begin{table}[ht]
\centering
\caption{The range of hyper-parameters for Ind-KNN.We apply the ViT extractor for CIFAR-10, the pre-trained ResNet50 model for Fashion MNIST and the sentence-transformer for AG News and DBpedia.} 
\begin{tabular}{c|ccccc}
    \toprule
   Hyper-parameters &  CIFAR-10& Fashion MNIST & AG News & DBpedia \\
    \hline
Noise scale $\sigma_1$ & \multicolumn{4}{c}{$\sqrt{\frac{T}{6B}}$}\\
\hline
  Noise scale $\sigma_2$ & \multicolumn{4}{c}{$\{0.1, 0.2, ..., 0.9\}$}\\
      \hline
 Minimum threshold $\tau$ (cosine kernel) &[0.10, 0.15]&[0.58, 0.63]&[0.35, 0.40]& [0.35, 0.40]\\
    \hline
 Minimum threshold $\tau$ (RBF kernel)& [0.68, 0.73]&  [0.8, 0.85]&- & -\\
   scale parameter $\nu$ (with RBF kernel) & $e^{1.5}$& $e^{1.5}$ &- &- \\
    \hline
\end{tabular}
%label{tab: time_vs_hash}
\end{table}
%\bibliography{egbib, DP}
% References
%\newpage
%appendix
%\onecolumn
%\input{ind_knn/10_appendix}

%% file: arxiv_main.bbl
\begin{thebibliography}{34}
\providecommand{\natexlab}[1]{#1}
\providecommand{\url}[1]{\texttt{#1}}
\expandafter\ifx\csname urlstyle\endcsname\relax
  \providecommand{\doi}[1]{doi: #1}\else
  \providecommand{\doi}{doi: \begingroup \urlstyle{rm}\Url}\fi

\bibitem[Abadi et~al.(2016)Abadi, Chu, Goodfellow, McMahan, Mironov, Talwar,
  and Zhang]{abadi2016deep}
Mart{\'\i}n Abadi, Andy Chu, Ian Goodfellow, H~Brendan McMahan, Ilya Mironov,
  Kunal Talwar, and Li~Zhang.
\newblock Deep learning with differential privacy.
\newblock In \emph{ACM SIGSAC Conference on Computer and Communications
  Security (CCS-16)}, pages 308--318. ACM, 2016.

\bibitem[Balle et~al.(2020)Balle, Barthe, Gaboardi, Hsu, and
  Sato]{balle2020hypothesis}
Borja Balle, Gilles Barthe, Marco Gaboardi, Justin Hsu, and Tetsuya Sato.
\newblock Hypothesis testing interpretations and renyi differential privacy.
\newblock In \emph{International Conference on Artificial Intelligence and
  Statistics}, pages 2496--2506. PMLR, 2020.

\bibitem[Bassily et~al.(2018)Bassily, Thakkar, and Guha~Thakurta]{bassily2018}
Raef Bassily, Om~Thakkar, and Abhradeep Guha~Thakurta.
\newblock Model-agnostic private learning.
\newblock \emph{Advances in Neural Information Processing Systems}, 31, 2018.

\bibitem[Bourtoule et~al.(2021)Bourtoule, Chandrasekaran, Choquette-Choo, Jia,
  Travers, Zhang, Lie, and Papernot]{bourtoule2021machine}
Lucas Bourtoule, Varun Chandrasekaran, Christopher~A Choquette-Choo, Hengrui
  Jia, Adelin Travers, Baiwu Zhang, David Lie, and Nicolas Papernot.
\newblock Machine unlearning.
\newblock In \emph{2021 IEEE Symposium on Security and Privacy (SP)}, pages
  141--159. IEEE, 2021.

\bibitem[Chaudhuri et~al.(2011)Chaudhuri, Monteleoni, and
  Sarwate]{chaudhuri2011differentially}
Kamalika Chaudhuri, Claire Monteleoni, and Anand~D Sarwate.
\newblock Differentially private empirical risk minimization.
\newblock \emph{The Journal of Machine Learning Research}, 12:\penalty0
  1069--1109, 2011.

\bibitem[Chen et~al.(2020)Chen, Kornblith, Norouzi, and Hinton]{clr}
Ting Chen, Simon Kornblith, Mohammad Norouzi, and Geoffrey Hinton.
\newblock A simple framework for contrastive learning of visual
  representations.
\newblock In \emph{International conference on machine learning}, pages
  1597--1607. PMLR, 2020.

\bibitem[Chen et~al.(2017)Chen, Liu, Li, Lu, and Song]{chen2017targeted}
Xinyun Chen, Chang Liu, Bo~Li, Kimberly Lu, and Dawn Song.
\newblock Targeted backdoor attacks on deep learning systems using data
  poisoning.
\newblock \emph{arXiv preprint arXiv:1712.05526}, 2017.

\bibitem[Dagan and Feldman(2020)]{dagan2020pac}
Yuval Dagan and Vitaly Feldman.
\newblock Pac learning with stable and private predictions.
\newblock In \emph{Conference on Learning Theory}, pages 1389--1410. PMLR,
  2020.

\bibitem[De et~al.(2022)De, Berrada, Hayes, Smith, and Balle]{de2022unlocking}
Soham De, Leonard Berrada, Jamie Hayes, Samuel~L Smith, and Borja Balle.
\newblock Unlocking high-accuracy differentially private image classification
  through scale.
\newblock \emph{arXiv preprint arXiv:2204.13650}, 2022.

\bibitem[Dosovitskiy et~al.(2020)Dosovitskiy, Beyer, Kolesnikov, Weissenborn,
  Zhai, Unterthiner, Dehghani, Minderer, Heigold, Gelly, et~al.]{vit}
Alexey Dosovitskiy, Lucas Beyer, Alexander Kolesnikov, Dirk Weissenborn,
  Xiaohua Zhai, Thomas Unterthiner, Mostafa Dehghani, Matthias Minderer, Georg
  Heigold, Sylvain Gelly, et~al.
\newblock An image is worth 16x16 words: Transformers for image recognition at
  scale.
\newblock \emph{arXiv preprint arXiv:2010.11929}, 2020.

\bibitem[Dwork and Feldman(2018)]{dwork2018privacy}
Cynthia Dwork and Vitaly Feldman.
\newblock Privacy-preserving prediction.
\newblock In \emph{Conference On Learning Theory}, pages 1693--1702. PMLR,
  2018.

\bibitem[Dwork et~al.(2006)Dwork, McSherry, Nissim, and
  Smith]{dwork2006calibrating}
Cynthia Dwork, Frank McSherry, Kobbi Nissim, and Adam Smith.
\newblock Calibrating noise to sensitivity in private data analysis.
\newblock In \emph{Theory of cryptography}, pages 265--284. Springer, 2006.

\bibitem[Dwork et~al.(2014)Dwork, Roth, et~al.]{dwork2014algorithmic}
Cynthia Dwork, Aaron Roth, et~al.
\newblock The algorithmic foundations of differential privacy.
\newblock \emph{Foundations and Trends{\textregistered} in Theoretical Computer
  Science}, 9\penalty0 (3--4):\penalty0 211--407, 2014.

\bibitem[Feldman and Zrnic(2021)]{feldman2021individual}
Vitaly Feldman and Tijana Zrnic.
\newblock Individual privacy accounting via a renyi filter.
\newblock \emph{Advances in Neural Information Processing Systems},
  34:\penalty0 28080--28091, 2021.

\bibitem[Gaboardi et~al.(2016)Gaboardi, Honaker, King, Nissim, Ullman, Vadhan,
  and Murtagh]{dptools}
Marco Gaboardi, James Honaker, Gary King, Kobbi Nissim, Jonathan Ullman, Salil
  Vadhan, and Jack Murtagh.
\newblock Psi ($\psi$): a private data sharing interface.
\newblock In \emph{Theory and Practice of Differential Privacy}, New York, NY,
  2016 2016.
\newblock URL \url{https://arxiv.org/abs/1609.04340}.

\bibitem[Ginart et~al.(2019)Ginart, Guan, Valiant, and Zou]{ginart2019making}
Antonio Ginart, Melody Guan, Gregory Valiant, and James~Y Zou.
\newblock Making ai forget you: Data deletion in machine learning.
\newblock \emph{Advances in neural information processing systems}, 32, 2019.

\bibitem[Gionis et~al.(1999)Gionis, Indyk, Motwani,
  et~al.]{gionis1999similarity}
Aristides Gionis, Piotr Indyk, Rajeev Motwani, et~al.
\newblock Similarity search in high dimensions via hashing.
\newblock In \emph{VLDB}, volume~99, pages 518--529, 1999.

\bibitem[Guo et~al.(2020)Guo, Goldstein, Hannun, and Van
  Der~Maaten]{guo2019certified}
Chuan Guo, Tom Goldstein, Awni Hannun, and Laurens Van Der~Maaten.
\newblock Certified data removal from machine learning models.
\newblock In \emph{Proceedings of the 37th International Conference on Machine
  Learning}, ICML'20. JMLR.org, 2020.

\bibitem[He et~al.(2016)He, Zhang, Ren, and Sun]{He_2016_CVPR}
Kaiming He, Xiangyu Zhang, Shaoqing Ren, and Jian Sun.
\newblock Deep residual learning for image recognition.
\newblock In \emph{Proceedings of the IEEE Conference on Computer Vision and
  Pattern Recognition (CVPR)}, 2016.

\bibitem[Jagielski et~al.(2018)Jagielski, Oprea, Biggio, Liu, Nita-Rotaru, and
  Li]{jagielski2018manipulating}
Matthew Jagielski, Alina Oprea, Battista Biggio, Chang Liu, Cristina
  Nita-Rotaru, and Bo~Li.
\newblock Manipulating machine learning: Poisoning attacks and countermeasures
  for regression learning.
\newblock In \emph{2018 IEEE Symposium on Security and Privacy (SP)}, pages
  19--35. IEEE, 2018.

\bibitem[Kasiviswanathan et~al.(2011)Kasiviswanathan, Lee, Nissim,
  Raskhodnikova, and Smith]{kasiviswanathan2011can}
Shiva~Prasad Kasiviswanathan, Homin~K Lee, Kobbi Nissim, Sofya Raskhodnikova,
  and Adam Smith.
\newblock What can we learn privately?
\newblock \emph{SIAM Journal on Computing}, 40\penalty0 (3):\penalty0 793--826,
  2011.

\bibitem[Krizhevsky et~al.(2009)Krizhevsky, Hinton,
  et~al.]{Krizhevsky2009LearningML}
Alex Krizhevsky, Geoffrey Hinton, et~al.
\newblock Learning multiple layers of features from tiny images.
\newblock \emph{Technical report}, 2009.

\bibitem[Lehmann et~al.(2015)Lehmann, Isele, Jakob, Jentzsch, Kontokostas,
  Mendes, Hellmann, Morsey, van Kleef, Auer, and Bizer]{Lehmann2015DBpediaA}
Jens Lehmann, Robert Isele, Max Jakob, Anja Jentzsch, Dimitris Kontokostas,
  Pablo~N. Mendes, Sebastian Hellmann, Mohamed Morsey, Patrick van Kleef,
  S.~Auer, and Christian Bizer.
\newblock Dbpedia - a large-scale, multilingual knowledge base extracted from
  wikipedia.
\newblock \emph{Semantic Web}, 6:\penalty0 167--195, 2015.

\bibitem[Mantelero(2013)]{mantelero2013eu}
Alessandro Mantelero.
\newblock The eu proposal for a general data protection regulation and the
  roots of the ‘right to be forgotten’.
\newblock \emph{Computer Law \& Security Review}, 29\penalty0 (3):\penalty0
  229--235, 2013.

\bibitem[Mironov(2017)]{mironov2017renyi}
Ilya Mironov.
\newblock R{\'e}nyi differential privacy.
\newblock In \emph{Computer Security Foundations Symposium (CSF), 2017 IEEE
  30th}, pages 263--275. IEEE, 2017.

\bibitem[Papernot et~al.(2018)Papernot, Song, Mironov, Raghunathan, Talwar, and
  Erlingsson]{pate2018}
Nicolas Papernot, Shuang Song, Ilya Mironov, Ananth Raghunathan, Kunal Talwar,
  and {\'U}lfar Erlingsson.
\newblock Scalable private learning with pate.
\newblock \emph{arXiv preprint arXiv:1802.08908}, 2018.

\bibitem[Reimers and Gurevych(2019)]{reimers-2019-sentence-bert}
Nils Reimers and Iryna Gurevych.
\newblock Sentence-bert: Sentence embeddings using siamese bert-networks.
\newblock In \emph{Proceedings of the 2019 Conference on Empirical Methods in
  Natural Language Processing}. Association for Computational Linguistics, 11
  2019.

\bibitem[Rogers et~al.(2016)Rogers, Roth, Ullman, and
  Vadhan]{rogers2016privacy}
Ryan~M Rogers, Aaron Roth, Jonathan Ullman, and Salil Vadhan.
\newblock Privacy odometers and filters: Pay-as-you-go composition.
\newblock \emph{Advances in Neural Information Processing Systems}, 29, 2016.

\bibitem[Tram{\`{e}}r and Boneh(2021)]{tramer2020differentially}
Florian Tram{\`{e}}r and Dan Boneh.
\newblock Differentially private learning needs better features (or much more
  data).
\newblock In \emph{9th International Conference on Learning Representations,
  {ICLR} 2021, Virtual Event, Austria, May 3-7, 2021}. OpenReview.net, 2021.
\newblock URL \url{https://openreview.net/forum?id=YTWGvpFOQD-}.

\bibitem[van~der Maaten and Hannun(2020)]{van2020trade}
Laurens van~der Maaten and Awni Hannun.
\newblock The trade-offs of private prediction.
\newblock \emph{arXiv preprint arXiv:2007.05089}, 2020.

\bibitem[Xiao et~al.(2017)Xiao, Rasul, and Vollgraf]{Xiao2017FashionMNISTAN}
Han Xiao, Kashif Rasul, and Roland Vollgraf.
\newblock Fashion-mnist: a novel image dataset for benchmarking machine
  learning algorithms.
\newblock \emph{ArXiv}, abs/1708.07747, 2017.

\bibitem[Zhang et~al.(2015)Zhang, Zhao, and LeCun]{Zhang2015CharacterlevelCN}
Xiang Zhang, Junbo~Jake Zhao, and Yann LeCun.
\newblock Character-level convolutional networks for text classification.
\newblock In \emph{Advances in Neural Information Processing Systems}, 2015.

\bibitem[Zhao et~al.(2022)Zhao, Yu, Wu, and Li]{zhao-etal-2022-compressing}
Xuandong Zhao, Zhiguo Yu, Ming Wu, and Lei Li.
\newblock Compressing sentence representation for semantic retrieval via
  homomorphic projective distillation.
\newblock In \emph{Findings of the Association for Computational Linguistics:
  ACL 2022}, pages 774--781, May 2022.

\bibitem[Zhu et~al.(2020)Zhu, Yu, Chandraker, and Wang]{zhu2020private}
Yuqing Zhu, Xiang Yu, Manmohan Chandraker, and Yu-Xiang Wang.
\newblock Private-knn: Practical differential privacy for computer vision.
\newblock In \emph{Proceedings of the IEEE/CVF Conference on Computer Vision
  and Pattern Recognition}, pages 11854--11862, 2020.

\end{thebibliography}
